\documentclass{article}
\usepackage[utf8]{inputenc}

\usepackage{footnote}
\usepackage{mathrsfs}
\usepackage{amsfonts}
\usepackage{multirow}
\usepackage{bm}
\usepackage{makecell}
\usepackage{amsmath}
\usepackage{graphicx}
\usepackage{algorithm,algorithmicx,algpseudocode}
\usepackage{caption}
\usepackage[numbers]{natbib}
\usepackage{geometry}
\usepackage{amsthm}

\usepackage{hyperref}
\usepackage[capitalise, noabbrev]{cleveref}

\usepackage{makeidx}
\usepackage{csquotes}
\usepackage{booktabs}
\usepackage{adjustbox}
\usepackage{comment}
\usepackage{color}
\usepackage{amsfonts}
\usepackage{titlesec}

\usepackage[affil-it]{authblk}
\usepackage[english]{babel}
\usepackage{blindtext}

\graphicspath{{Figures/}}

\newcommand{\R}{{\mathbb R}}
\newcommand{\N}{{\mathbb N}}

\newcommand{\E}{{\mathbb E}}

\newcommand{\cP}{{\mathcal P}}
\newcommand{\cT}{{\mathcal T}}
\newcommand{\cQ}{{\mathcal Q}}

\newcommand{\dd}{\ifmmode\bm{d}\else\textbf{\textit{d}}\fi}
\newcommand{\w}{\ifmmode\bm{w}\else\textbf{\textit{w}}\fi}
\newcommand{\W}{\ifmmode\bm{W}\else\textbf{\textit{W}}\fi}
\newcommand{\X}{\ifmmode\bm{X}\else\textbf{\textit{X}}\fi}
\newcommand{\x}{\ifmmode\bm{x}\else\textbf{\textit{x}}\fi}
\newcommand{\y}{\ifmmode\bm{y}\else\textbf{\textit{y}}\fi}
\newcommand{\bb}{\ifmmode\bm{b}\else\textbf{\textit{b}}\fi}
\newcommand{\z}{\ifmmode\bm{z}\else\textbf{\textit{z}}\fi}
\newcommand{\ttt}{\ifmmode\bm{t}\else\textbf{\textit{t}}\fi}
\newcommand{\uu}{\ifmmode\bm{u}\else\textbf{\textit{u}}\fi}
\newcommand{\vv}{\ifmmode\bm{v}\else\textbf{\textit{v}}\fi}
\newcommand{\UU}{\ifmmode\bm{U}\else\textbf{\textit{U}}\fi}
\newcommand{\VV}{\ifmmode\bm{V}\else\textbf{\textit{V}}\fi}
\newcommand{\m}{\ifmmode\bm{m}\else\textbf{\textit{m}}\fi}
\newcommand{\M}{\ifmmode\bm{M}\else\textbf{\textit{M}}\fi}
\newcommand{\cc}{\ifmmode\bm{c}\else\textbf{\textit{c}}\fi}
\newcommand{\nn}{\ifmmode\bm{n}\else\textbf{\textit{n}}\fi}
\title{Semi Conditional Variational Auto-Encoder for Flow Reconstruction and Uncertainty Quantification from Limited Observations}

\author[1,*]{Kristian Gundersen}
\author[1]{Anna Oleynik}
\author[2]{Nello Blaser}
\author[1]{Guttorm Alendal}
\affil[1]{Department of Mathematics, University of Bergen}
\affil[2]{Department of Informatics, University of Bergen}

\date{\today}

\begin{document}

\maketitle
\begin{abstract}
We present a new data-driven model to reconstruct nonlinear flow from spatially sparse observations. The model is a version of a conditional variational auto-encoder (CVAE), which allows for probabilistic reconstruction and thus uncertainty quantification of the prediction. We show that in our model, conditioning on the measurements from the complete flow data leads to a CVAE where only the decoder depends on the measurements. For this reason we call the model as Semi-Conditional Variational Autoencoder (SCVAE). The method, reconstructions and associated uncertainty estimates are illustrated on the velocity data from simulations of 2D flow around a cylinder and bottom currents from the Bergen Ocean Model. The reconstruction errors are compared to those of the Gappy Proper Orthogonal Decomposition (GPOD) method. 
\end{abstract}

\section{Introduction}
Reconstruction of non-linear dynamic processes based on sparse observations is an important and difficult problem. The problem traditionally requires knowledge of the governing equations or processes to be able to generalize from the the sparse observations to a wider area around, in-between and beyond the measurements. Alternatively it is possible to learn the underlying processes or equations based on data itself, so called data driven methods. In geophysics and environmental monitoring measurements is often only available at sparse locations. For instance, within the field of meteorology, atmospheric pressures, temperatures and wind are only measured at limited number of stations. To produce accurate and general weather predictions, requires methods that both forecast in the future, but also reconstruct where no data is available. Within oceanography one faces the same problem, that in-situ information about the ocean dynamics is only available at sparse locations such as buoys or sub-sea sensors. \\

Both the weather and ocean currents can be approximated with models that are governed by physical laws, e.g. the Navier-Stokes Equation. However, to get accurate reliable reconstructions and forecasts it is of crucial importance to incorporate observations. \\ 

Reconstruction and inference based on sparse observations is important in many applications both in engineering and physical science \cite{brunton2015closed, kong2018application, bolton2019applications, venturi2004gappy,callaham2018robust, manohar2018data}. Bolton et. al. \cite{bolton2019applications}  used convolutional neural networks to hindcast ocean models, and in \cite{yeo2019data} K. Yeo reconstructs time series of nonlinear dynamics from sparse observation.  Oikonomo et. al. \cite{oikonomou2018novel} proposed a method for filling data gaps in groundwater level observations and Kong. et. al \cite{kong2018application} used reconstruction techniques to modeling the characteristics of cartridge valves.  \\

The above mentioned applications are just some of the many examples of reconstruction of a dynamic process based on limited information. Here we focus on reconstruction of flow. This problem can be formulated as follows. Let $\bm w \in \R^d,$ $d \in \N,$ represent a state of the flow, for example velocity, pressure, temperature, etc. Here, we will focus on incompressible unsteady flows and ${\bm w}=(u,v)\in \R^2$ where $u$ and $v$ are the horizontal and vertical velocities, respectively. The velocities ${\bm w}$ are typically obtained from computational fluid dynamic simulations on a meshed spatial domain $\mathcal{P}$ at discrete times $\mathcal{T} = \{t_1,...,t_K\} \subset \R$. \\

Let $ \mathcal{P}=\{p_1,...,p_N\}$ consist of $N$ grid points $p_n,$ $n=1,...,N.$  
Then the state of the flow $\bm w$ evaluated on $\cP$ at a time $t_i \in \cT$ can be represented as a vector 
$\x^{(i)} \in \R^{2N},$ 
\begin{equation}
\label{eq:x_i}
\x^{(i)}=(u(p_1,t_i),...,u(p_N,t_i), v(p_1,t_i),...,v(p_N,t_i))^T.
\end{equation}
The collection of $\x^{(i)},$ $i=1,\dots,K,$ constitutes the data set $\X.$ In order to account for incompressibility, we introduce a discrete divergence operator $L_{div}$, which is given by a $N \times 2N$ matrix associated with a finite difference scheme, and
\begin{equation}
\label{eq:L_div}
(L_{div} \, \x)_{k} \approx (\nabla \cdot w)(p_k)=0.
\end{equation}

\begin{figure}[h]
	\centering
		\includegraphics[width=0.70\linewidth]{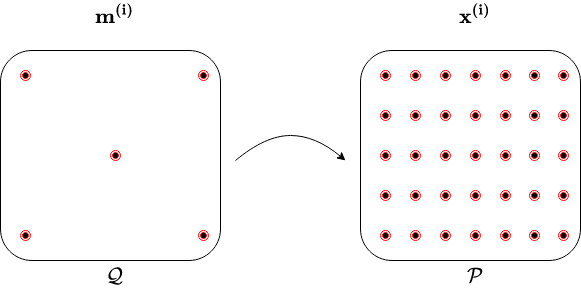}
		\captionof{figure}{Sketch of reconstruction of $\x^{(i)}$ from $\m^{(i)}$. The dots on the right side represent the grid $\cP$, and those on  the left side represent the measurement locations $\cQ.$}\label{Sketch_map}
\end{figure}

Further, we assume that the state can be measured only at specific points in $\cP,$ that is, at $\cQ=\{q_1,...,q_M\} \subset \cP$ where $M$ is typically much less than $N.$  Hence, there is $\M=\{\m^{(i)} \in \R^{2M}: \, \m^{(i)}=C\,\x^{(i)}, \, \forall \x^{(i)} \in \X \},$ where ${\bm C} \in \R^{2M \times 2N}$ is a sampling matrix. More specifically, $\bm C$ is a two block matrix 

$$
{\bm C}=\begin{pmatrix}
{\bm C}_{1/2}& O\\
O&{\bm C}_{1/2}\\
\end{pmatrix}, \quad 
({\bm C}_{1/2})_{ij}= \left\{
\begin{array}{ll}
1,& \mbox{if } \, q_i =p_j\\
0,& \mbox{otherwise}
\end{array} \right. , \quad i=1,...,N \quad j=1,...,M,
$$
and ${\bm O}\in \R^{M\times N}$ is a zero matrix. The problem of reconstructing fluid flow $\x^{(i)} \in \X$ from $\m^{(i)}\in \M$ is presented as a schematic plot in \cref{Sketch_map}.
There have been a wide range of methods for solving the problem, e.g. \cite{sirovich1987turbulence, everson1995karhunen, donoho2006compressed, schmid2010dynamic, ELMS2018, raissi2019physics}. In particular, use of proper orthogonal decomposition (POD) \cite{sirovich1987turbulence} techniques has been popular. \\ 

POD \cite{sirovich1987turbulence} is a traditional dimensional reduction technique where based on a data set, a number of basis functions are constructed. The key idea is that a linear combination of the basis functions can reconstruct the original data within some error margin, efficiently reducing the dimension of the problem. In a modified version of the POD, the Gappy POD (GPOD) \citep{everson1995karhunen}, the aim is to fill the gap in-between sparse measurements. Given a POD basis one can minimize the $L_2$-error of the measurements and find a linear combination of the POD-basis that complements between the measurements. If the basis is not know, a iterative scheme can be formulated to optimize the basis based on the measurements. The original application of GPOD \citep{everson1995karhunen} was related to reconstruction of human faces, and it has later been applied to fluid flow reconstruction \cite{venturi2004gappy}. We will use the GPOD approach for comparison later in this study. \\ 

A similar approach is the technique of Compressed Sensing (CS) \cite{donoho2006compressed}. As for the GPOD method, we want to solve a  linear system. However, in the CS-case this will be a under-determined linear system. That is we need some additional information about the system to be able to solve it, typically this can be a condition/constraint related to the smoothness of the solution. The core difference between CS and GPOD is however the sparsity constraint. That is, instead of minimizing the L2-norm, we minimize the L1-norm. Minimizing the L1-norm favours sparse solutions, i.e. solutions with a small number of nonzero coefficients. \\ 

Another reconstruction approach is Dynamical Mode Decomposition (DMD) \cite{schmid2010dynamic}. Instead of using principal components in the spatial domain, DMD seek to find modes or representations that are associated with a specific frequency in the data, i.e. modes in the temporal domain. Again, the goal is to find a solution to an undetermined linear system and reconstruct based on the measurements, by minimizing the error of the observed values. \\

During the last decade, data driven methods have become tremendously popular, partly because of the growth and availability of data, but also driven by new technology and improved hardware. To model a non-linear relationships with linear approximations is one of the fundamental limitation of the DMD, CS and GPOD. Recently we have seen development in methods where the artificial neural networks is informed with a physical law, the so called physic-informed neural networks (PINN) \cite{raissi2019physics}. In PINNs the reconstruction is informed by a Partial Differential Equation (PDE) (e.g. Navier Stokes), thus the neural network can learn to fill the gap between measurements that are in compliance with the equation. This is what Rassi et. al. \cite{raissi2018hidden} have shown for the benchmark examples such as flow around a 2D and 3D cylinder. Although PINNs are showing promising results, we have yet to see applications to complex systems such as atmospheric or oceanographic systems, where other aspect have to be accounted for, e.g. in large scale oceanic circulation models that are driven by forcing such as tides, bathymetry and river-influx. That being said, these problems may be resolved through PINNs in the future. Despite the promise of PINNs, they will not be a part of this study, as our approach is without any constraint related to the physical properties of the data. \\

Another non-linear data driven approaches for reconstruction of fluid flow are different variations of auto-encoders \cite{ELMS2018, grover2019uncertainty}. An auto-encoder \cite{Rumelhart86_autoencoder} is a special configuration of an artificial neural network that first encodes the data by gradually decreasing the size of the hidden layers. With this process, the data is represented in a lower dimensional space. A second neural network then takes the output of the encoder as input, and decodes the representation back to its original shape. These two neural networks together constitute an auto-encoder. Principal Component Analysis (PCA) \cite{pearson1901_PCA} also represent the data in a different and more compact space. However, PCA reduce the dimension of the data by finding orthogonal basis functions or principal components through singular value decomposition. In fact, it has been showed with linear activation function, PCA and auto-encoders produces the same basis function \cite{bourlard1988auto}. Probabilistic version of the auto-encoder are called Variational Auto-Encoders (VAEs) \cite{kingma2013auto}. CVAEs \cite{sohn2015learning} are conditional probabilistic auto-encoders, that is, the model is dependent on some additional information such that it is possible to create representations that are depend on this information. \\

Here, we address the mentioned problem from a probabilistic point of view. Let $\x: \cP \to \R^{2N}$ and $\m: \cQ \to \R^{2M}$ be two multivariate random variables associated with the flow on $\cP$ and on $\cQ$, respectively. Then the data sets $\X$ and $\M$ consist of the realizations of $\x$ and $\m$, respectively. Using $\X$ and $\M,$ we intend to approximate the probability distribution $p(\x|\m).$ This would not only allow to predict $\x^{(i)}$ given $\m^{(i)},$ but also to estimate an associated uncertainty. In this paper, we use a variational auto-encoder to approximate $p(\x| \m)$. The method we use is a Bayesian Neural Network \cite{MacKay92} approximated through variational inference \cite{hoffman2013stochastic,blei2017variational}, that we have called \textit{Semi-Conditional Variational Auto-encoder}, SCVAE. A detailed description of the SCVAE method for reconstruction and associated uncertainty quantification is given in \cref{SCVAE_section}. \\

Here we focus on fluid flow, being the main driving mechanism behind transport and dilution of tracers in marine waters. The world's oceans are  under tremendous stress \citep{Halpern:2012hs}, UN has declared 2021-2030 as the ocean decade\footnote{\url{https://en.unesco.org/ocean-decade}}, and an ecosystem based Marine Spatial Planning initiative has been launched by IOC \citep{DominguezTejo:2016dt}. \\

Local and regional current conditions determines transport of tracers in the ocean \cite{drange2001ocean,BARSTOW1983211}. Examples are accidental release of radioactive, biological or chemical substances from industrial complexes, e.g. organic waste from fish farms in Norwegian fjords \citep{Ali:2011hd}, plastic \cite{Law:2017}, or other contaminants that might have adverse effects on marine ecosystems \citep{Hylland:2015gt}. \\ 

To be able to predict the environmental impact of a release, i.e. concentrations as function of distance and direction from the source, requires reliable current conditions \citep{Ali:2016go,Blackford:2020}. Subsequently, these transport predictions support design of marine environmental monitoring programs \citep{Hvidevold:2015,Hvidevold:2016cx,Alendal:2017b, oleynik2020optimal}. The aim here is to model current conditions in a probabilistic manner using SCVAEs. This allows for predicting footprints in a Monte Carlo framework, providing simulated data for training networks used for, e.g., analysing environmental time series \cite{gundersen2020binary}.\\ 

In this study we will compare results  with the GPOD method \cite{willcox2006unsteady}. We are aware that there recent methods (e.g. PINNS and traditional Auto-encoder) that may perform better on the specific data sets than the GPOD, however, the GPODs simplicity, versatility and not least its popularity \cite{jo2019effective, mifsud2019fusing, callaham2019robust}, makes it a great method for comparison. \\

The reminder of this manuscript is outlined in the following: \cref{A_Motivating_Example} presents a motivating example for the SCVAE-method in comparison with the GPOD-method. In \cref{methods} we review both the VAE and CVAE method and present the SCVAE. Results of experiments on two different data sets are presented in \cref{Experiment}. \cref{discussion} summarize and discuss the method, experiments, drawbacks and benefits and potential extensions and further work.  

\section{A Motivating Example}\label{A_Motivating_Example}
Here we illustrate the performance of the proposed method vs the GPOD method in order to give a motivation for this study. We use simulations of a two dimensional viscous flow around a cylinder at the Raynolds number of $160,$ obtained from \url{https://www.csc.kth.se/~weinkauf/notes/cylinder2d.html}. The simulations were performed  by Weinkauf et. al. \cite{weinkauf2010streak} with the Gerris Flow Solver software \cite{gerrisflowsolver}. The data set consists of a horizontal $u$ and a vertical $v$ velocities on an uniform $400 \times 50 \times 1001$ grid of $[-0.5, 7.5] \times [-0.5, 0.5] \times [15, 23]$ spatial-temporal domain.\footnote{The simulations are run from $t=0$ to $t=23$, but velocities are only extracted from $t=15$ to $t=23$} In particular, we have $400$ points in the horizontal, and $50$ points in the vertical  direction, and $1001$ points in time.  
\begin{figure}[h]
	\centering
		\includegraphics[width=0.99\linewidth]{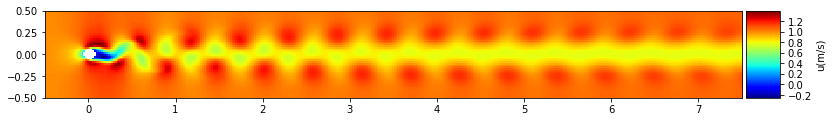}
    	\includegraphics[width=0.99\linewidth]{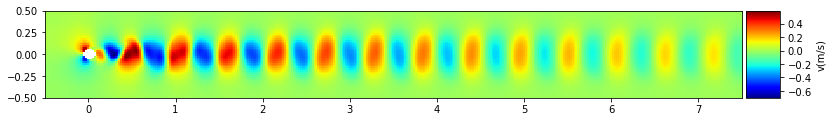}
		\captionof{figure}{Typical data instance from the original 2D flow around a cylinder data set with $u$ and $v$ component presented at the upper and lower panel, respectively} \label{Cw_data_motivating}
\end{figure}
The cylinder has the diameter of $0.125$ and is centered at the origin, see \cref{Cw_data_motivating}. The left vertical boundary (inlet) has Dirichlet boundary condition $u=1$ and $v=0$. The homogeneous Neumann boundary condition is given at the right boundary (outlet), and the homogeneous Dirichlet conditions on the remaining boundaries. At the start of simulations, $t=0$, both velocities were equal to zero.  We plot the velocities at the time $t \approx 19$ (time step $500$) in \cref{Cw_data_motivating}. \\

For simplicity, in the experiment below we extract the data downstream from the cylinder, that is, from grid point $40$ to $200$ in the horizontal direction, and keep all grid points in vertical direction. Hence, $\mathcal{P}$ contains $N = 8000$ points, $160$ points in the horizontal and $50$ in the vertical direction. The temporal resolution is kept as before, that is, the number of time steps in $\mathcal{T}$ is $K=1001$. For validation purposes, the data set was split into a train, validation and test data set. The train and validation data sets were used for optimization of the model parameters. For both the SCVAE and the GPOD, the goal was to minimize the $L2$ error between the true and the modeled flow state.  
The restriction of the GPOD is that the number of components $r$ could be at most $2M.$ 
To deal with this problem, and to account for the flow incompressibility, we added the  regularization term  $\lambda \|L_{div} x^{(i)}\|,$ $\lambda>0$, to the objective function, see \cref{Appendix_C}. For the GPOD method, the parameters $r$ and/or $\lambda$ where optimized on the validation data set in order to have the smallest mean error.
We give more details about objective functions for the SCVAE in \cref{SCVAE_section}. For now we mention that there are two versions, where one version uses an additional divergence regularization term similar to GPOD.\\

\begin{figure}[h!]
    \centering
	\includegraphics[width=0.75\linewidth]{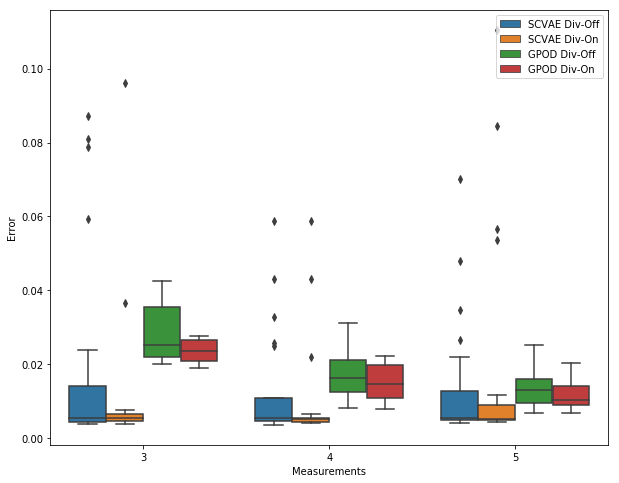}
		\captionof{figure}{The mean relative error for two reconstruction methods. 
		The orange and blue label correspond to the SCVAE with (div-on) and without (div-off) additional divergence regularization. The green and red labels correspond to the GPOD method. }\label{Error_boxplot_CW}
\end{figure}

In \cref{Error_boxplot_CW} we plot the mean  of the relative $L_2$ error calculated on the test data for both methods with and without the div-regularization. The results are presented for $3,$ $4,$ and $5$ measurement locations, that is, $M=3,4,5.$ For each of these three cases, we selected $20$ different configurations of $M.$. In particular, we created $20$ subgrids $\cQ$, each containing $5$ randomly sampled spatial grid points. Next we removed one and then two points from each of the $20$ subgrids $\cQ,$ to create new subgrids of $4$ and $3$ measurements, respectively.
As it can be seen in \cref{Error_boxplot_CW}, both methods perform well for the $5$ measurements case. The resulting relative errors have comparable mean and variance. When reducing the number of observations, the SCVAE method maintains low errors, while the GPOD error increases. The SCVAE seems to benefit from the additional regularization of minimizing the divergence, in terms of lower error and less variation in the error estimates. The effect is more profound with fewer measurements. \\

The key benefit of the SCVAE is that its predictions are optimal for the given measurement locations.  In a contrast, the POD based approaches, and in particular the GPOD, create a set of basis functions (principal components) based on the training data independently of the measurements. While this has an obvious computational advantage, the number of principle components for complex flows can be high and, as a result, many more measurements are needed, \cite{willcox2006unsteady,manohar2018data,Proctor2014}. There are number of algorithms that aim to optimize to measurement locations to achieve the best performance of the POD based methods, see e.g., \cite{jo2019effective,willcox2006unsteady,YILDIRIM2009160}. In practice, however, the locations are often fixed and another approaches are needed. The results in \cref{Error_boxplot_CW} suggest that the SCVAE could be one of these approaches.

\section{Methods}\label{methods}
Before we introduce the model used for reconstruction of flows, we give a brief introduction to VAEs and CVAEs. For a detailed introduction, see \cite{vae_intro}. VAEs are neural network models that has been used for learning structured representations in a  wide variety of applications, e.g., image generation \cite{gregor2015draw}, interpolation between sentences \cite{bowman2015generating} and compressed sensing \cite{grover2019uncertainty}. 

\subsection{Preliminary}
Let us assume that the data $\X$ is generated by a random process that involves an unobserved continuous random variable $\z.$ The process consists of two steps: (i) a value $\z^{(i)}$ is sampled from a prior $p_{\theta^*}(\z);$ and (ii)  $\x^{(i)}$ is generated from a conditional distribution  $p_{\theta^*}(\x|\z).$ In the case of flow reconstruction, $\z$ could be thought of as unknown boundary or initial conditions, tidal and wind forcing, etc. However, generally $\z$ is just a convenient construct to represent $\X,$ rather than a physically explained phenomena. Therefore it is for convenience assumed that $p_{\theta^*}(\z)$ and $p_{\theta^*}(\x|\z)$ come from parametric families of distributions $p_{\theta}(\z)$ and $p_{\theta}(\x|\z),$ and their density functions are differentiable almost everywhere w.r.t. both $\z$ and $\theta$. A probabilistic auto-encoder is neural network that is trained to represent its input $\X$ as $p_\theta(\x)$ via \textit{latent representation} $\z \sim p_{\theta}(\z),$ that is,

\begin{equation} \label{eq:p_theta(x)}
p_{\theta}(\x) = \int p_{\theta}(\x,\z) d\z = \int p_{\theta}(\x|\z)p_{\theta}(\z)d\z.
\end{equation}
As $p_\theta(\z)$ is unknown and observations $\z^{(i)}$ are not accessible, we must use $\X$ in order to generate $\z \sim p_\theta(\z|\x).$ That is, the network can be viewed as consisting of two parts: an \textit{encoder} $p_\theta(\z|\x)$ and a \textit{decoder} $p_\theta(\x|\z).$ Typically the true posterior distribution $p_{\theta}(\z|\x)$ is intractable but could be approximated with variational inference \cite{hoffman2013stochastic,blei2017variational}.  That is, we define a so called recognition model $q_{\phi}(\z|\x)$ with variational parameters $\phi$, which aims to approximate $p_{\theta}(\z|\x).$ The recognition model is often parameterized as a Gaussian. Thus, the problem of estimating $p_{\theta}(\z|\x)$, is reduced to finding the best possible estimate for $\phi$, effectively turning the problem into an optimization problem. \\

An auto-encoder that uses a recognition model is called Variational Auto-Encoder (VAE). In order to get good prediction we need to estimate the parameters $\phi$ and $\theta.$ The marginal likelihood  is equal to the sum over the marginal likelihoods of the individual samples, that is, $\sum_{i=1}^K \log p_{\theta}(\x^{(i)}).$ Therefore, we further on present estimates for an individual sample. The Kullback -Leibler divergence between two probability distributions  $q_{\phi}(\z|\x^{(i)})$ and $p_{\theta}(\z|\x^{(i)})$, defined as $$D_{KL}[q_{\phi}(\z|\x^{(i)})||p_{\theta}(\z|\x^{(i)})] = \int q_{\phi}(\z|\x^{(i)}) \log\left(\frac{q_{\phi}(\z|\x^{(i)})}{p_{\theta}(\z|\x^{(i)})}\right) d\z,$$ can be interpreted as a measure of distinctiveness between these two distributions \cite{kullback1951information}. It can be shown, see \cite{vae_intro}, that
\begin{equation}\label{eq:DKL_via_L}
\log p_{\theta}(\x^{(i)})=D_{KL}[q_{\phi}(\z|\x^{(i)})||p_{\theta}(\z|\x^{(i)})] +\mathcal{L}(\theta,\phi;\x^{(i)}),
\end{equation}
where
$$
\mathcal{L}(\theta,\phi;\x^{(i)}) = 
\E_{q_{\phi}(\z|\x^{(i)})} \left[ -\log q_{\phi}(\z|\x^{(i)})+\log p_{\theta}(\x^{(i)},\z) \right].
$$
Since KL-divergence is non-negative, we have $\log p_{\theta}(\x^{(i)}) \geq\mathcal{L}(\theta,\phi;\x^{(i)})$ and
$\mathcal{L}(\theta,\phi;\x^{(i)})$
is called Evidence Lower Bound (ELBO) for the marginal likelihood $\log p_{\theta}(\x^{(i)}).$
Thus, instead of maximizing the marginal probability, one can instead maximize its variational lower bound to which we also refer as an objective function. It can be further shown that the ELBO can be written as
\begin{equation}
	\label{eq:VLB:2_no_beta}
	\mathcal{L}(\theta,\phi;\x^{(i)})= \E_{q_{\phi}(\z|\x^{(i)})} \left[ \log p_{\theta}(\x^{(i)}|\z) \right] - D_{KL}[q_{\phi}(\z|\x^{(i)})||p_{\theta}(\z)].
\end{equation}
Reformulating the traditional VAE framework as a constraint optimization problem, it is possible to obtain the $\beta$-VAE \cite{higgins2016beta} objective function if $p_\theta(\z) =\mathcal{N}({\bf 0},{\bm I}),$
\begin{equation}
	\label{eq:VLB:2}
	\mathcal{L}(\theta,\phi;\x^{(i)})= \E_{q_{\phi}(\z|\x^{(i)})} \left[ \log p_{\theta}(\x^{(i)}|\z) \right] - \beta D_{KL}[q_{\phi}(\z|\x^{(i)})||p_{\theta}(\z)],
\end{equation}
where $\beta>0.$ Here $\beta$ is a regularisation coefficient that constrains the capacity of the latent representation $\z$. The $\E_{q_{\phi}(\z|\x^{(i)})} \left[ \log p_{\theta}(\x^{(i)}|\z) \right]$ can be interpreted as the reconstruction term, while the KL-term, $\beta D_{KL}[q_{\phi}(\z|\x^{(i)})||p_{\theta}(\z)]$ as regularization term.

Conditional Variational Auto-encoders \cite{sohn2015learning} (CVAE) are similar to VAEs, but differ by conditioning on an additional property of the data (e.g. a label or class), here denoted $\cc$. Conditioning both the recognition model and the true posteriori on both $\x^{(i)}$ and $\cc$ results in the CVAE ELBO 
\begin{align}
    \begin{split}
        \mathcal{L}(\theta,\phi;\x^{(i)},\cc)=\E_{q_{\phi}(\z|\x^{(i)}, \cc)} \left[ \log p_{\theta}(\x^{(i)}|\z,\cc) \right] - D_{KL}[q(\z|\x^{(i)},\cc)||p_{\theta}(\z|\cc)]. \label{CVAE_objective}
    \end{split}
\end{align}
In the decoding phase, CVAE allows for conditional probabilistic reconstruction and permits sampling from the conditional distribution $p_{\theta}(\z|\cc)$, which has been useful for generative modeling of data with known labels, see \cite{sohn2015learning}. Here we investigate a special case of the CVAE when $\cc$ is a partial observation of $\x.$ We call this Semi Conditional Variational Auto-encoder (SCVAE).

\subsection{Semi Conditional Variational Auto-encoder}\label{SCVAE_section}
\begin{figure}[ht]
    \centering
		\includegraphics[width=0.80\linewidth]{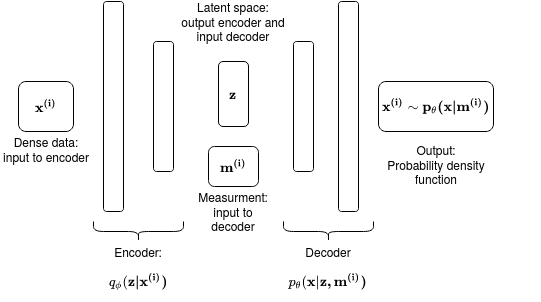}
		\captionof{figure}{The figure shows a sketch of the model used to estimate 
		$p_{\theta}(\x|\m^{(i)})$. During training both the observations $\m^{(i)}$ and the data $\x^{(i)}$ will be used. After the model is trained, we can predict using only the decoder part of the neural network. The input to the decoder will then only be the observations and random samples from the latent space.}\label{DAE_neural_network_sketch}
\end{figure}

The SCVAE takes the input data $\X$, conditioned on $\M$ and approximates the probability distribution $p_{\theta}(\x|\z,\m).$ Then we can generate $\x^{(i)}$, based on the observations $\m^{(i)}$ and latent representation $\z$. As $\m^{(i)}=C \x^{(i)}$ where $C$ is a non-stochastic sampling matrix, we have 
$$p_{\theta}(\z|\x^{(i)}, \m^{(i)}) = p_{\theta}(\z|\x^{(i)}), \, \mbox{ and } \, q_{\phi}(\z|\x^{(i)},\m^{(i)})=q_{\phi}(\z|\x^{(i)}).$$ Therefore, from \cref{CVAE_objective} the ELBO for SCVAE is 
\begin{equation}
    \begin{split}
        \log p_{\theta}(\x^{(i)}|\m^{(i)})\geq \mathcal{L}(\theta,\phi;\x^{(i)},\m^{(i)})=
        &\E_{q_{\phi}(\z|\x^{(i)})} \left[ \log p_{\theta}(\x^{(i)}|\z, \m^{(i)}) \right] \\ & - D_{KL}[q_\phi(\z|\x^{(i)})||p_{\theta}(\z|\m^{(i)})]\label{eq:ELBO:SCVAE_no_beta}
    \end{split}
\end{equation}
where $p_\theta(\z|\m^{(i)}) = \mathcal{N}({\bf 0},{\bf I}).$ Similarly as for the $\beta$-VAE \cite{higgins2016beta} we can obtain a relaxed version of \cref{eq:ELBO:SCVAE_no_beta} by maximizing the parameters $\{\phi, \theta\}$ of the expected log-likelihood $\E_{q_\phi(\cdot)} [\log p_\theta (\x^{(i)}|\m^{(i)},\z)])$ and treat it as an constrained optimization problem. That is, 
\begin{align}
    \begin{split}
    & \max\limits_{\phi,\theta} \E_{q_\phi(\cdot)} [\log p_\theta (\x^{(i)}|\m^{(i)},\z)])  \text{ subject to} \\ &
    D_{KL}(q_\phi(\z|\m^{(i)},\x^{(i)})||p_\theta (\z|\m^{(i)}) \leq \epsilon  
    \end{split}\label{beta_SCVAE_opt_prob}
\end{align}
where $\epsilon>0$ is small. The subscript $q_\phi(\cdot)$ is short for $q_\phi(\z|\m^{(i)},\x^{(i)}).$ Since $\m^{(i)}$ is dependent on $\x^{(i)}$ we have that $q_\phi(\z|\m^{(i)},\x^{(i)}) = q_\phi(\z|\x^{(i)}).$ \cref{beta_SCVAE_opt_prob}  can expressed as a Lagrangian under the Karush–Kuhn–Tucker (KKT) conditions \cite{kuhn2014nonlinear, karush1939minima}. Hence,
\begin{align}
    \begin{split}
    \mathcal{F}(\theta, \phi, \beta, \alpha, \x^{(i)}, \m^{(i)}) & = 
    \E_{q_\phi(\cdot)} [\log p_\theta (\x^{(i)}|\m^{(i)},\z)]) \\ & + 
    \beta(D_{KL}(q_\phi(\z|\x^{(i)})||p_\theta (\z|\m^{(i)})-\epsilon)   
    \end{split}\label{beta_SCVAE_constrained}
\end{align}
According to the complementary slackness KKT condition $\beta \geq 0,$ we can rewrite \cref{beta_SCVAE_constrained} as
\begin{align}
    \centering
    \begin{split}
      \mathcal{F}(\theta, \phi, \beta, \x^{(i)}, \m^{(i)}) \geq  \mathcal{L}(\theta, \phi, \x^{(i)}, \m^{(i)}) & =  \E_{q_\phi(\cdot)} [\log p_\theta (\x^{(i)}|\m^{(i)},\z)])  \\ & +
     \beta D_{KL}(q_\phi(\z|\x^{(i)})||p_\theta (\z|\m^{(i)}).   
    \end{split}\label{beta_SCVAE_constrained_2}
\end{align}
Objective functions in \cref{eq:ELBO:SCVAE_no_beta} and \cref{beta_SCVAE_constrained_2}, and later \cref{eq:ELBO:SCVAE:J}, show that if conditioning on a feature which is a known function of the original data, such as measurements, we do not need to account for them in the encoding phase.The measurements are then coupled with the encoded data in the decoder. We sketch the main components of the SCVAE in \cref{DAE_neural_network_sketch}.
In order to preserve some physical properties of the data $\X,$ we can condition yet on another feature. Here we utilize the incompressibility property of the fluid, i.e., $\dd^{(i)}=L_{div} \x^{(i)} \approx 0,$ see \cref{eq:L_div}. \\ 

We intend to maximize a log-likelihood under an additional constrain $\dd^{(i)}$, compared to \cref{beta_SCVAE_opt_prob}. That is
\begin{align}
    \begin{split}
    & \max\limits_{\phi,\theta} \E_{q_\phi(\cdot)} [\log p_\theta (\x^{(i)}|\m^{(i)},\z)])  \text{ subject to} \\ &
    D_{KL}(q_\phi(\z|\x^{(i)})||p_\theta (\z|\m^{(i)},\dd^{(i)}) \leq \epsilon \quad \text{and} \quad \\ &
    -\E_{q_\phi(\cdot)}[\log p_\theta(\dd^{(i)}|\m^{(i)},\z)] \leq \delta 
    \end{split}\label{Constraint_optimization_prob_v3}
\end{align}
where $\epsilon, \delta>0$ are small. \cref{Constraint_optimization_prob_v3}  can expressed as a Lagrangian under the Karush–Kuhn–Tucker (KKT) conditions as before and as a consequence of the complementary slackness condition $\lambda,\beta \geq 0,$ we can obtain the objective function 

\begin{align}
    \centering
    \begin{split}
      \mathcal{F}(\theta, \phi, \beta, \alpha, \x^{(i)}, \m^{(i)}, \dd^{(i)}) \geq  \mathcal{L}(\theta, \phi, \x^{(i)}, \m^{(i)}, \dd^{(i)}) & =  \E_{q_\phi(\cdot)} [\log p_\theta (\x^{(i)}|\m^{(i)},\z)])  \\ & +
     \lambda\,\E_{q_\phi(\cdot)} [\log p_\theta (\dd^{(i)}|\m^{(i)},\z)])  \\ & -
     \beta D_{KL}(q_\phi(\z|\x^{(i)})||p_\theta (\z|\m^{(i)},\dd^{(i)}),   
    \end{split}\label{eq:ELBO:SCVAE:J}
\end{align}
where $p(\z|\m^{(i)}, \dd^{(i)}) = \mathcal{N}({\bf 0},{\bf I}).$ For convenience of notation we refer to the objective function \cref{beta_SCVAE_constrained_2} as the case with $\lambda=0$, and the objective function \cref{eq:ELBO:SCVAE:J} as the case with $\lambda > 0.$ Observe that under the Gaussian assumptions on the priors, \cref{eq:ELBO:SCVAE:J} is equivalent to \cref{beta_SCVAE_constrained_2} if $\lambda=0.$ Thus, from now one we will refer to it as a special case of \cref{eq:ELBO:SCVAE:J} and denote as $\mathcal{L}_{0}.$ \\

Similarly to \cite{kingma2013auto} we obtain $q_{\phi}(\z|\x^{(i)})= \mathcal{N}({\mu}^{(i)} \mathbf{1},(\sigma^{(i)})^2\,\mathbf{I})
$, that is, $\phi=\{ \mu, \sigma\}.$  This allows to express the KL-divergence terms in a closed form and avoid issues related to differentiability of the ELBOs.  Under these assumptions, the KL-divergence terms can be integrated analytically while the term
$\E_{q_{\phi}(\z|\x^{(i)})} \left[ \log p_{\theta}(\x^{(i)}|\z, \m^{(i)}) \right] $ and $\E_{q_{\phi}(\z|\x^{(i)})} \left[ \log p_{\theta}(\dd^{(i)}|\z, \m^{(i)}) \right] $
requires estimation by sampling
\begin{equation}\label{eq:E-estimate}
    \begin{array}{l}
         \E_{q_{\phi}(\z|\x^{(i)})} \left[ \log p_{\theta}(\x^{(i)}|\z, \m^{(i)}) \right] \approx 
        \frac{1}{L}\sum\limits_{l=1}^{L} \log  p_{\theta}(\x^{(i)}|\z^{(i,l)},\m^{(i)}),\\
        \E_{q_{\phi}(\z|\x^{(i)})} \left[ \log p_{\theta}(\dd^{(i)}|\z, \m^{(i)}) \right] \approx 
        \frac{1}{L}\sum\limits_{l=1}^{L} \log  p_{\theta}(\dd^{(i)}|\z^{(i,l)},\m^{(i)}),\\
        \mbox{where } \, \z^{(i,l)} = g_{\phi}(\bm{\epsilon}^{(i,l)}, \x^{(i)}), \quad 
        \bm{\epsilon}^{l} \sim p(\bm{\epsilon}).
    \end{array}
\end{equation}
Here $\bm{\epsilon}^{l}$ is an auxiliary (noise) variable with independent marginal $p(\bm{\epsilon})$, and $g_{\phi}(\cdot)$ is a differentiable transformation of $\bm{\epsilon},$ parametrized by $\phi,$  see for details \cite{kingma2013auto}. We denote $\mathcal{L}_\lambda,$ $\lambda \geq 0$  \cref{eq:ELBO:SCVAE:J} with the approximation above as $\mathcal{\widehat{L}_\lambda},$ that is,
\begin{align}
    \begin{split}
        & \widehat{\mathcal{L}}_\lambda(\theta, \phi, \x^{(i)},\m^{(i)},\dd^{(i)})  =   \frac{1}{L}\sum\limits_{l=1}^{L} \log  p_{\theta}(\x^{(i)}|\z^{(i,l)},\m^{(i)}) \\
         &+ \lambda \frac{1}{L}\sum\limits_{l=1}^{L} \log p_{\theta}(\dd^{(i)}|\z^{(i,l)},\m^{(i)}) -\beta D_{KL}[q_{\phi}(\z|\x^{(i)})||p_{\theta}(\z|\m^{(i)}, \dd^{(i)})].
        \label{Loss_function}
    \end{split}
\end{align}
The objective function $\widehat{\mathcal{L}}_\lambda$ can be maximized by gradient descent. Since the gradient $\nabla_{\theta,\phi}\,\widehat{\mathcal{L}}_\lambda$ cannot be calculated for large data sets, Stochastic Gradient Descent methods, see \cite{kiefer1952stochastic, robbins1951stochastic} are typically used where 
\begin{equation}
    \widehat{\mathcal{L}}_\lambda(\theta, \phi; \X, \M, \bm{D}) \approx \widehat{\mathcal{L}}^{R}(\theta, \phi; \X^R, \M^R, \bm{D}^R) = 
    \frac{K}{R}\sum\limits_{r=1}^{R} \widehat{\mathcal{L}}_\lambda(\theta, \phi; \x^{(i_r)}, \m^{(i_r)}, \dd^{(i_r)}), \quad \lambda \geq 0.\label{obj_function_SCVAE}
\end{equation}
Here  $\X^{R}=\left\{\x^{(i_r)}\right\}_{r=1}^{R},$  $R<K$ is a minibatch consisting of randomly sampled datapoints, $\M^{R}=\left\{\m^{(i_r)}\right\}_{r=1}^{R}$ and $\bm{D}^{R}=\left\{\dd^{(i_r)}\right\}_{r=1}^{R}.$ After the network is optimized, a posterior predictive distribution $p_\theta(\x|\m, \dd)$ can be approximated with a Monte Carlo estimator. \\

\subsubsection{Uncertainty Quantification}\label{sec:posterior}
Let  $\hat{\theta}$ and $\hat{\phi}$ be an estimation of generative and variational parameters, as described in \cref{SCVAE_section}. Then the decoder can be used to predict the posterior as
\begin{align}\label{eq:p_pred}
    p_{\hat{\theta}}(\x|\m^*, \dd^*) \approx \frac{1}{N_{MC}} \sum_{j=1}^{N_{MC}} p_{\hat{\theta}}(\x|\z^{(j)},\m^*, \dd^*) \xrightarrow[N_{MC} \rightarrow \infty]{} \int
    p_{\hat{\theta}}(\x|\z, \m^*, \dd^*)p_{\hat{\theta}}(\z|\m^*,\dd^*)d\z.
\end{align}
While sampling from the latent space has been viewed typically as an approach for generating new samples with similar properties, here we use it to estimate the prediction uncertainty of the trained model. From \cref{eq:p_pred} we are able to estimate the mean prediction  $\hat{\x}^*$ and empirical covarience matrix $\widehat{{\bm \Sigma}}$ using  a Monte Carlo estimator. We get
\begin{align} 
      \widehat{\x}^*=
      \frac{1}{N_{MC}} \sum\limits_{j=1}^{N_{MC}} \x^{(j)}\quad \mbox{and} \quad
      \widehat{{\bm \Sigma}} =  \frac{1}{N_{MC}-1} \sum\limits_{j=1}^{N_{MC}}(\x^{(j)} -  \widehat{\x}^*)(\x^{(j)} -  \widehat{\x}^*)^T,
        \label{eq:mean_and_cov} 
 \end{align}
where $\x^{(j)}  \sim p_{\hat{\theta}}(\x|\m^*,\dd^*).$ The empirical standard deviation is then simply $\widehat{\bm \sigma} = \sqrt{diag(\widehat{{\bm \Sigma}})}.$ To estimate the confidence region we assume that the predicted $p_{\hat{\theta}}(\x|\m^{*},\dd^*)$ is well approximated by a normal distribution $N({\bm \mu}, \bm{\Sigma}).$ Given that $\widehat{\x}^*$ and $\widehat{{\bm \Sigma}}$ are approximations of $\bm{\mu}$ and $\bm{\Sigma},$ obtained from $N_{MC}$ samples as above, a confidence region estimate for a prediction $\x^{*}$ can be given as
\begin{equation}\label{eq:ConfRegion:1}
    \left\{\x^{(i)} \in \R^{2N}: \,(\x^{(i)}-{\widehat{\x}^*})^T \hat{\bm \Sigma}^{+} (\x^{(i)}- \widehat{\x}^*)\leq \chi^2_k(p)\right\}
\end{equation}
where $\chi^2_k(p)$ is the quantile function for probability $p$ of the chi-squared distribution with $k=\min\{N_{MC},2N\}$ degrees of freedom, and $\widehat{\bm \Sigma}^{+}$ is the pseudoinverse of $\widehat{\bm \Sigma}.$ Using the singular value decomposition, ${\widehat{\bm \Sigma}}=\bm{U S U}^T,$ the corresponding interval for $(\x^*)_n,$ $n=1,\dots,2N,$ is
\begin{equation}\label{eq:ConfRegion:2}
   \left[
 ({\widehat{\x}^*})_n-\sqrt{\chi^2_k(p)}\,\|{\bm u}_n^T\,{\bm S}^{1/2}\|_2,\quad
 ({\widehat{\x}^*})_n+\sqrt{\chi^2_k(p)}\,
 \|{\bm u}_n^T\,{\bm S}^{1/2}\|_2
  \right]   
\end{equation}
where ${\bm u}_n^T$ is $n$th row of the matrix ${\bm U}.$

\section{Experiments}\label{Experiment}
We will present the SCVAE method on two different data sets. The first one is the 2D flow around a cylinder described in \cref{A_Motivating_Example}, and the second is ocean currents on the seafloor created by the Bergen Ocean Model \cite{berntsen2000users}. The data $\X$ consists of the two dimensional velocities ${\bm w} =(u, v).$ To illustrate the results we will plot $u$ and $v$ components of $\x^{(i)} \in \X,$ see \cref{eq:x_i}. For validation of the models, the data $\X$ is split into train, test and validation subsets, which are subscripted accordingly, if necessary. The data sets, spitting and preprocessing for each case are described in \cref{Cylinder_wake_experiment,BOM_experiment}. \\

We use a schematically simple architecture to explore the SCVAE. The main ingredient of the encoder is convolutional neural network (CNN)  \cite{lecun1998gradient, krizhevsky2012imagenet} and for the decoder we use transposed CNN-layers \cite{noh2015learning}. The SCVAE has a slightly different architecture in each case, which we present in \cref{Appendix_A}. \\

The SCVAE is trained to maximize the objective functions in \cref{obj_function_SCVAE} with the backpropagation algorithm \cite{lecun1989backpropagation} and the Adam algorithm \cite{kingma2014adam}. We use an adaptive approach of weighing the reconstruction term with KL-divergence and/or divergence terms \cite{heydari2019softadapt}, that is, finding the regularization parameters $\beta$ and $\lambda$. Specifically, we calculate the proportion of each term contribution of the total value of the objective function, and scale the terms accordingly. This approach prevents posterior collapse. Posteriori collapse occurs if the KL-divergence term becomes too close to zero, resulting in a non probabilistic reconstruction. The approach of weighing the terms proportionally iteratively adjusts the weight of the KL-divergence term, $\beta$, such that posterior collapse is avoided. For the result shown here we trained the SCVAEs  with an early stopping criteria of $50$ epochs, i.e., the optimization is stopped if we do not see any improvement after $50$ epochs, and returns the best model. We use a two-dimensional Gaussian  distribution for $p_\theta(\z|\m^{(i)}, \dd^{(i)})$ in all the experiments.\\

Let the test data $\X_{test}$ consist of $n$ instances $\x^{(i)},$ $i=1,\dots,n,$ and $\widehat{\x}^{(i)}$ denote a prediction of the true $\x^{(i)}$ given $\m^{(i)}.$ In the case of the SCVAE, $\widehat{\x}^{(i)}$ is the mean prediction obtained as in \cref{eq:mean_and_cov}. For the GPOD method, $\widehat{\x}^{(i)}$ is a deterministic output of the optimization problem, see \cref{Appendix_C}. In order to compare the SCVAE results with the results of the GPOD method, we introduce the following measures; the mean of the relative error for the prediction
\begin{equation}\label{L2_error}
  \mathcal{E} = \frac{1}{n}\sum\limits_{i=1}^{n}\frac{||\widehat{\x}^{(i)} - \x^{(i)}||_2}{||\x^{(i)}||_2}
\end{equation}
and the mean of the absolute error for the divergence 
\begin{equation}
        \mathcal{E}_{div} = \frac{1}{n}\sum\limits_{i=1}^{n}\| L_{div} \, \x^{(i)}\|_2.\label{divergence_error_1}
\end{equation}

\subsection{2D Flow Around a Cylinder}\label{Cylinder_wake_experiment}
Here we return to the example in \cref{A_Motivating_Example}. Below we give some additional details of the data preprocessing and model implementation.
\subsubsection{Preprocessing}
The data is reduced as described in \cref{A_Motivating_Example}. 
We assess the SCVAE with a sequential split for train, test and validation. The sequantial split is defined as follows. The last $15 \%$ of the data is used for test, the last $30\%$ of the remaining data is used for validation, and the first $70\%$ for training. To improve the conditioning of the optimization problem we scale the data as decribed in  \cref{Appendix_D}. The errors $\mathcal{E}$ (\cref{L2_error}) and $\mathcal{E}_{div} $ (\cref{divergence_error_1}) are calculated after scaling the data back. The input to the SVAE $\x^{(i)}$ was shaped as $(160 \times 50 \times 2)$ in order to apply convolutional layers. Here we use  $5,4,3$ and $2$ fixed spatial measurements, that is, four different subgrids $\cQ$ 
\begin{align}
    \centering
    \begin{split}
    \cQ_5 = & \{(12,76),(47,8),(30,40),(153,34),(16,10)\}, \\
    \cQ_4 = & \{(12,76),(47,8),(30,40),(153,34)\},  \\
    \cQ_3 = & \{(12,76),(47,8),(30,40)\}, \\
    \cQ_2 = & \{(12,76),(47,8)\}.
    \end{split}\label{CW_measurements}
\end{align}
The flow state at these specific locations constitute $\M$.

\subsubsection{Model}
A schematic description of the model is given in \cref{Appendix_A}. 
The first layer of the encoder is a zero-padding layer that expands the horizontal and vertical dimension by adding zeros on the boundaries. Here we add zero-padding of four in the horizontal and three in the vertical direction. The subsequent layers consists of two convolutional layers, where the first and second layer have $160$ and $200$ filters, respectively. We use a kernel size and strides of $2$ in both convolutional layers and ReLu activation functions. This design compresses the data into a $(42 \times 14 \times 200)$ shape. The compressed representation from the convolutional layers are flattened and are further compressed into a $64$ dimensional vector through a traditional dense layer. Two outputs layers are defined to represent the mean and log-variance of the latent representation $\z$. The reparametrization trick is realized in a third layer, a so called  lambda layer, which takes the mean and log-variance as an input and generates $\z$. The output of the encoder are the samples $\z^{(i)}$ and the mean and the log-variance of $\z$. \\

The decoder takes the latent representation $\z^{(i)}$ and the measurements $\m^{(i)}$ as input. The input $\m^{(i)}$ is flattened and then concatenated with $\z^{(i)}$. The next layer is a dense layer with shape $(42 \times 14 \times 200)$. Afterwards there are two transposed convolutional layers with filters of $200$ and $160$. The strides and the kernel size is the same as for the encoder. The final layer is a transposed convolutional layer, with same dimension as the input to the encoder, that is the dimension of $\x^{(i)}$. A linear activation function is used for this output layer. The last layer of the model is a lambda layer that removes the zero-padding. In the next section we show statistics of the probabilistic reconstruction and compare with the GPOD method. 

\subsubsection{Results}
\begin{figure}[H]
	\centering
		\includegraphics[width=0.99\linewidth]{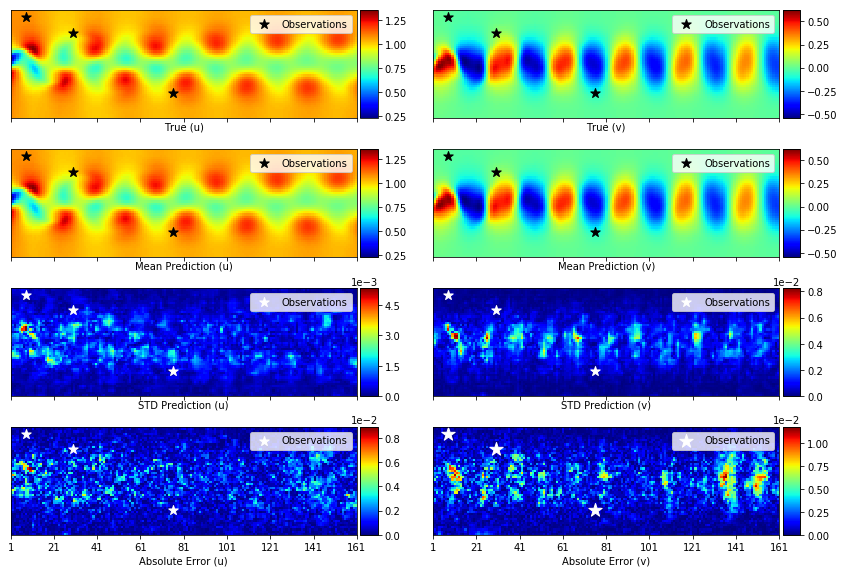}
	\captionof{figure}{Left panels shows the u-velocities, and the right panel v-velocities. The results are based on a model trained with $\lambda=0$ and $Q_3$ measurement locations. \textbf{First panels:} The true solutions \textbf{Second panels:} Reconstructed solution based on the SCVAE model \textbf{Third panels:} Standard deviation of the predicted solution \textbf{Fourth panels:} Absolute error between the true and predicted solution.}\label{Cylinder_wake_pred}
\end{figure}

In \cref{Cylinder_wake_pred} we have plotted the true velocity field, the reconstructed velocities, the standard deviation of the velocities and the absolute error between the true and reconstructed velocity fields. The observations placement is shown as stars (black and white). The SCVAE with the objective function  $\widehat{\mathcal{L}}_0$, see \cref{CVAE_objective}, was used for this prediction. To generate the posterior predictive distributions, \cref{eq:p_pred},  we sample $100$ realizations from $\z \sim \mathcal{N}(\bm{0},\bm{I})$ , which allows for calculation mean prediction and uncertainty estimates, see \cref{eq:mean_and_cov}. \\

We emphasise here again that the SCVAE with $\widehat{\mathcal{L}}_0$ and with $\widehat{\mathcal{L}}_\lambda,$ $\lambda>0,$ are two different models. For the notation sake we here refer to $\lambda=0$ when we mean the model with the objective function in \cref{CVAE_objective}, and to $\lambda>0$ when in \cref{eq:ELBO:SCVAE:J}. The same holds for the GPOD method, see \cref{Appendix_D}. When $\lambda=0,$ the number of the principle components $r$ is less $2M.$ The number $r$ is chosen such that the prediction on the validation data has the smallest possible error on average. If $\lambda>0,$ no restrictions on $r$ are imposed. In this case both  $\lambda$ and $r$ are estimated from the validation data. \\

The general observation is that the SCVAE reconstruction fits the data well, with associated low uncertainty. 
This can be explained by the periodicity in the data. In particular, the training and validation data sets represent the  test data well enough.

\begin{figure}[H]
	\centering
		\includegraphics[width=0.85\linewidth]{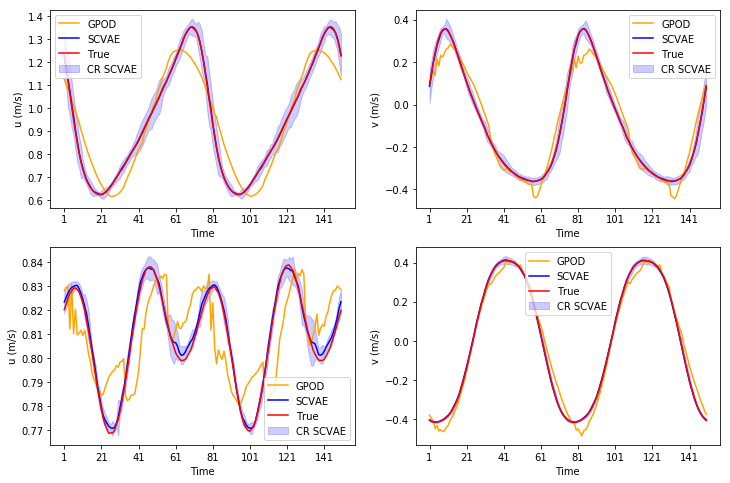}
		\captionof{figure}{Velocities $u$ and $v$ at specific locations. The red line corresponds to the true values, blue to the SCVAE mean prediction, and orange to the GPOD reconstruction. Light blue shaded area represents the confidence region obtained in \cref{eq:ConfRegion:2} with $p=0.95.$ The results are obtained from the model trained with $\lambda=0$ and $Q_3$ measurement locations. \textbf{Upper panels:} The time series and confidence region for $u$ (left) and $v$ (right) at  grid point $(6, 31).$ \textbf{Lower panels:} The time series and confidence region for $u$ (left) and $v$ (right) at grid point $(101, 25).$ }\label{Cylinder_wake_time_series}
\end{figure}

In \cref{Cylinder_wake_time_series} we  have plotted four time series of the reconstructed test data at two specific grid points, together with the confidence regions constructed as in \cref{eq:ConfRegion:2} with $p=0.95.$ The two upper panels represents the reconstruction at the grid point $(6, 31)$, and the lower at $(101,25)$ for $u$ and $v$ on the left and right side, respectively. The SCVAE reconstruction is significantly better than the GPOD, and close to the true solution for all time steps. 
\begin{figure}[H]
	\centering
		\includegraphics[width=0.85\linewidth]{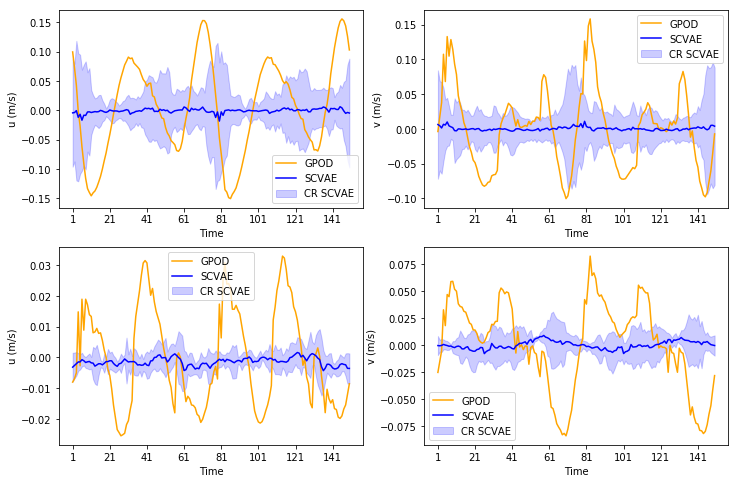}
		\captionof{figure}{The difference between the true and predicted estimate for the SCVAE (blue) and for the GPOD (orange). The light blue shaded region represents the difference marginals, obtained from the confidence region in \cref{BOM_time_series}. The estimates are based on a model trained with $\lambda=0$ and $Q_3$ measurement locations. \textbf{Upper panels:} The difference between the true and predicted estimate at grid point $(6,31)$ for $u$ (left) and $v$ (right), \textbf{Lower panels:} The difference between the true and predicted estimate at point $(101,25)$ for $u$ (left) and $v$ (right).}\label{Cylinder_wake_error}
\end{figure}
\cref{Cylinder_wake_error} shows the difference between the true values and the model prediction in time for the same two locations. This figure has to be seen in context with \cref{Cylinder_wake_pred}. In \cref{tab:comparison_CW}  we display the relative errors, \cref{L2_error}, for the SCVAE and the GPOD method, both with and without divergence regularization, for $5, 4, 3,$ and $2$ measurement locations given in \cref{CW_measurements}. \\

The results of the SCVAE depend on two stochastic inputs which are (i) randomness in the initialization of the prior weights and (ii) random mini batch sampling. We have trained the model with a each measurement configuration $10$ times, and chose the model that performs the best on the validation data set. Ideally we would run test cases where we used all the values as measurements,i.e., $\M=\X,$ and test how well the model would reconstruct in this case. This would then give us the lower bound of the best reconstruction that is possible for this specific architecture and hyper parameter settings. However, this scenario was not possible to test, due to limitations in memory in the GPU. Therefore we have used a large enough $M$ which still allowed us to run the model. In particular, we used every fifth and second pixel in the horizontal and vertical direction, which resulted in a total of $(32 \times 25)$ measurement locations, or $M=800$. We believe that training the model with these settings, gave us a good indication of the lower bound of the reconstruction error. The error observed was of the magnitude of $10^{-3}$. \\

\begin{figure}[H]
    \centering
 	    \includegraphics[width=0.75\linewidth]{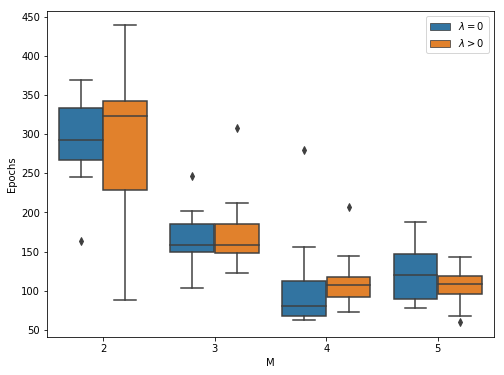}
 	    \captionof{figure}{Number of epochs trained depending on the number of measurements. For each measurement configuration and regularization technique the model is run $10$ times. The variation of number of epochs for for each measurement locations is due to different priors of the weights and random mini-batch sampling.}\label{Epochs_per_measurement_CW}
\end{figure}

This lower bound has been reached for all measurement configurations \cref{CW_measurements}. 
However, larger computational cost was needed to reach the lower bound for fewer measurement locations. \cref{Epochs_per_measurement_CW} shows the number of epochs as a boxplot diagram. In comparison with GPOD, the SCVAE error is 10 times lower than the GPOD error, and this difference becomes larger with fewer measurements. Note that adding regularization did not have much effect on  the relative error. From the motivating example we observed that regularizing with $\lambda>0$ is better in terms of a more consistent and low variable error estimation. Here we selected from the 10 trained models the one that performed best on the validation data set. This model selection approach shows that there are no significant differences between the two regularization techniques. The associated error in the divergence of the velocity fields is reported in  \cref{tab:comparison_CW_div}. 
\begin{table}[H]
    \centering
    \begin{tabular}{|c|c|c|c|c|c|} 
        \hline
         \multirow{2}{*}{Method} & \multirow{2}{*}{Regularization} & \multicolumn{4}{|c|}{Measurement Locations} \\  \cline{3-6} 
          & & 5 & 4 & 3 & 2  \\  \hline
         \multirow{2}{*}{SCVAE} & $\lambda = 0$
         & 0.30e-02 & 0.33e-02 & 0.26e-02 & 0.28e-02    \\ \cline{2-6}
         & $\lambda > 0$  & 0.31e-02 & 0.32e-02 & 0.30e-02 & 0.28e-02 \\ \cline{3-6} \hline
         \multirow{2}{*}{GPOD} & $\lambda = 0$  & 2.35e-02& 2.49e-02  
           & 3.38e-02& 17.38e-02\\ \cline{2-6}
         & $\lambda > 0$ &   2.12e-02 & 2.33e-02& 3.15e-02& 16.38e-02    \\ \hline
    \end{tabular}
    \caption{The mean relative error $\mathcal{E}$ (\cref{L2_error}) for the  SCVAE prediction  and  the GPOD prediction with or without div-regularization, and different number of measurements.}
    \label{tab:comparison_CW}
\end{table}
\begin{table}[H]
    \centering
    \begin{tabular}{|c|c|c|c|c|c|} 
        \hline
         \multirow{2}{*}{Method} & \multirow{2}{*}{Regularization} & \multicolumn{4}{|c|}{Measurement Locations} \\  \cline{3-6} 
          & & 5 & 4 & 3 & 2  \\  \hline
         
         \multirow{2}{*}{SCVAE} & $\lambda = 0$
          & 0.1439 & 0.1580 & 0.1383 & 0.1432    \\ \cline{2-6}
         & $\lambda > 0$ & 0.1533 & 0.1408 & 0.1468 & 0.1410    \\ \cline{3-6} \hline
         \multirow{2}{*}{GPOD} & $\lambda = 0$ & 0.1052 & 0.1047 & 0.0943 & 0.08866    \\ \cline{2-6}
         & $\lambda > 0$  & 0.1039 & 0.1051 & 0.0966 & 0.0669   \\ \hline
    \end{tabular}
    \caption{Comparison of the divergence error $\mathcal{E}_{div}$ as calculated in \cref{divergence_error_1} for the different methods and regularization techniques. The true divergence error on the entire test data set is $0.1058$}
    \label{tab:comparison_CW_div}
\end{table}

\subsection{Current data from Bergen ocean model}\label{BOM_experiment}
We tested the SCVAE on simulations from the Bergen Ocean Model (BOM) \cite{berntsen2000users}. BOM is a three-dimensional terrain-following nonhydrostatic ocean model with capabilities of resolving mesoscale to large-scale processes. Here we use velocities simulated by Ali. et. al \cite{Ali:2016go}. The simulations where conducted on the entire North Sea with 800 meter horizontal and vertical grid resolution and 41 layers for the period from 1st to 15th of January 2012. Forcing of the model consist of wind, atmospheric pressure, harmonic tides, rivers, and initial fields for salinity and temperature. For details of the setup of the model, forcing and the simulations we refer to \cite{Ali:2016go}. \\

\begin{figure}[H]
	\centering
		\includegraphics[width=0.99\linewidth]{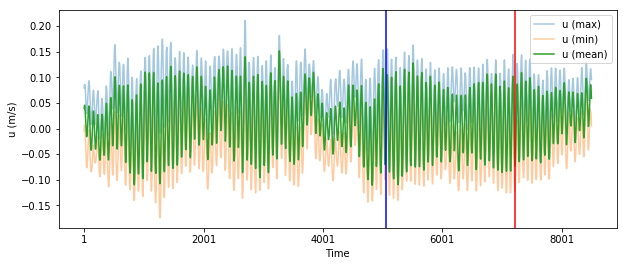}
        \includegraphics[width=0.99\linewidth]{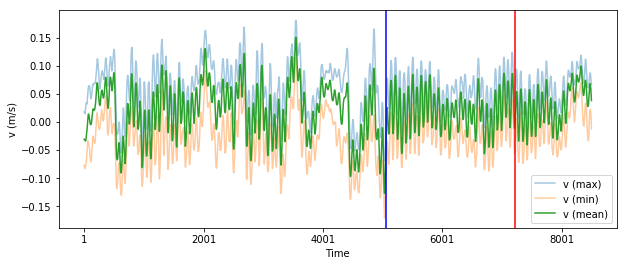}
		\captionof{figure}{The light-blue line represent the maximum, the orange the minimum and the green mean value of $u$ and $v$ for each time $t$ in $\cT.$ The horizontal lines indicate the sequential data split. }\label{Data_set_time_series_BOM}
\end{figure}

Here, the horizontal and vertical velocities of an excerpt of 25.6 $\times$ 25.6 km$^2$ at the bottom layer centered at the Sleipner CO2 injection site ($58.36^\circ N, \, 1.91^\circ E$) is used as data set for reconstruction. In \cref{Data_set_time_series_BOM} we have plotted the mean and extreme values of $u$ and $v$ for each time $t$ in $\cT$. 
\subsubsection{Preprocessing}
We extract $32 \times 32$ central grid from the bottom layer velocity data. Hence, $\mathcal{P}$ contains $N = 1024$ points, $32$ points in the horizontal and $32$ in the vertical direction. The temporal resolution is originally $105000$ and the time between each time step is $1$ minute. We downsample the temporal dimension of the original data uniformly such that the number of time steps in $\mathcal{T}$ is $K=8500.$ We train and validate the SCVAE with two different data splits: randomized and sequential in time. For the sequential split we have used the last $15 \%$ for the test, the last $30\%$ of the remaining data is used for validation, and the fist $70\%$ for training. In \cref{Data_set_time_series_BOM}, the red and blue vertical lines indicate the data split for this case. For the random split, the instances $\x^{(i)}$ are drawn randomly from $\X$ with the same percentage. The data was scaled as described in \cref{Appendix_D}.  The input $\x^{(i)}$ to the SCVAE was shaped as $(32 \times 32 \times 2)$ in order to apply convolutional layers. We use  $9,5$ and $3$ fixed spatial measurement locations. In particular, the subgrid $\cQ$ is given as 
\begin{align}\label{BOM_measurements}
    \centering
    \begin{split}
    \cQ_9 = & \{(6,6),(6,17),(6,27),(17,17),(17,27),(17,6),(27,6), (27,17),(27,27) \}, \\
    \cQ_5 = & \{(6,6),(17,17),(27,27),(6,27),(27,6) \}, \\
    \cQ_3 = & \{ (6,27),(17,17),(27,6) \}. 
    \end{split}
\end{align}
As before, the values of $u$ and $v$ at these specific locations constitute the measurements $\m^{(i)} \in \M$.

\subsubsection{Model} 
A schematic description of the model is given in \cref{Appendix_A.3,Appendix_A.4}. 
The first two layers of the encoder are convolutional layers with $64$ and $128$ filters with strides and kernel size of $2$ and ReLu activation functions. This compresses the data into a shape of $(8 \times 8 \times 128)$. The next layers are flattening and dense layers, where the latter have $16$ filters and ReLu activation. The subsequent layers defines the mean and log-variance of the latent representation $\z$, which is input to a lambda layer for realization of the reparametrization trick. The encoder outputs the samples $\z^{(i)}$ and the mean and the log-variance of $\z$.\\

Input to the decoder is the output $\z^{(i)}$ of the encoder and the measurement $\m^{(i)}.$ To concatenate the inputs, $\m^{(i)}$ is flattened. After concatenation of $\z^{(i)}$ and $\m^{(i)}$, the next layer is a dense layer with shape $(8 \times 8 \times 128)$ and ReLu activation. This allows for use of transposed convolutional layers to obtain the original shape of the data. Hence, the following layers are two transposed convolutional layers with $64$ and $128$ filters, strides and kernel size of $2$ and ReLu activation's. The final layer is a transposed convolutional with linear activation functions and filter size  of shape $(32 \times 32 \times 2),$ i.e., the same shape as $\x^{(i)}$. 

\subsubsection{Results}\label{BOM_results}
\begin{figure}[H]
    \centering
		\includegraphics[width=0.75\linewidth]{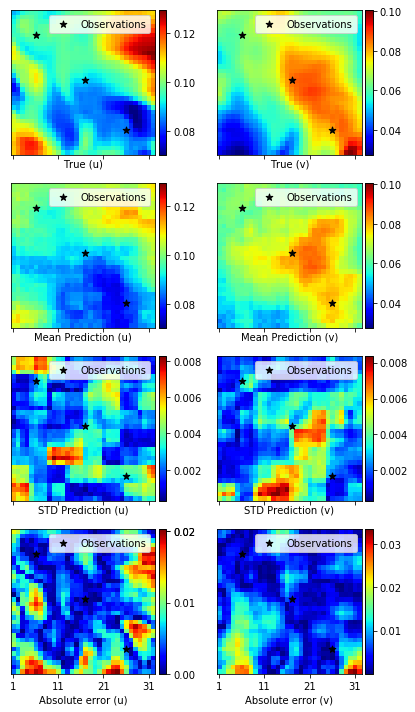}
		\captionof{figure}{Presentation of statistics for
		the reconstruction of the $u$ and $v$ component of the velocity for sample $1185$ in the test data set based on the trained model with $\lambda=0$ and $Q_3$ measurement locations. \textbf{Left panels:} From top to bottom; True velocity in $u$, predicted mean velocity field of $u$, the standard deviation of the prediction for $u$ and the absolute error of $u.$ \textbf{Lower panels:} Similar as describe for the upper panels, but for $v$ }\label{UV_prediction}
\end{figure}

We illustrate the obtained posterior predictive distribution in terms the predictive mean and standard deviation for the prediction at a specific time. The SCVAE is compared with the GPOD method, both with $\lambda >0$ and $\lambda = 0$ for measurement locations given in \cref{BOM_measurements} for random and sequential split cases. To generate the posterior predictive distributions, \cref{eq:p_pred},  we sample $200$ realizations from $\z \sim \mathcal{N}(\bm{0},\bm{I})$ , which allows for calculation mean prediction and uncertainty
estimates, see \cref{eq:mean_and_cov}. \cref{UV_prediction} shows the results of the prediction at time step $1185$ for both the $u$ and $v$ component and associated uncertainty estimates for a trained model with $\lambda=0$ and $Q_3$ measurement locations (see \cref{BOM_measurements}).  

\begin{figure}[H]
        \centering
		\includegraphics[width=0.85\linewidth]{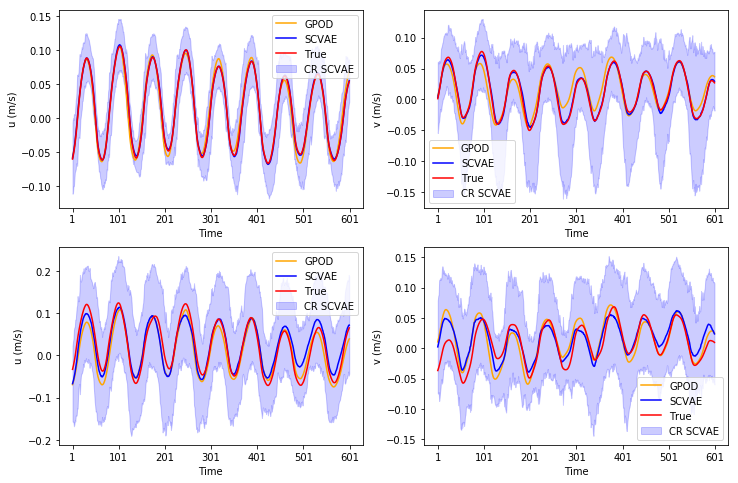}
		\captionof{figure}{Velocities $u$ and $v$ at specific locations based on the trained model with $\lambda=0$ and $Q_3$ measurement locations. The red line corresponds to the true values, blue to the SCVAE mean prediction, and orange to the GPOD reconstruction. Light blue shaded area represents the confidence region obtained in \cref{eq:ConfRegion:2} with $p=0.95.$.  \textbf{Upper panels:} The time series and confidence region for $u$ (left) and $v$ (right) at  grid point $(26,6)$, approximately $5.1$ km from nearest observation. \textbf{Lower panels:} The time series and confidence region for $u$ (left) and $v$ (right) at grid point $(4,1)$ approximately $16.1$ km from nearest observation.}\label{BOM_time_series}
\end{figure}

\begin{figure}[H]
	\centering
	\includegraphics[width=0.85\linewidth]{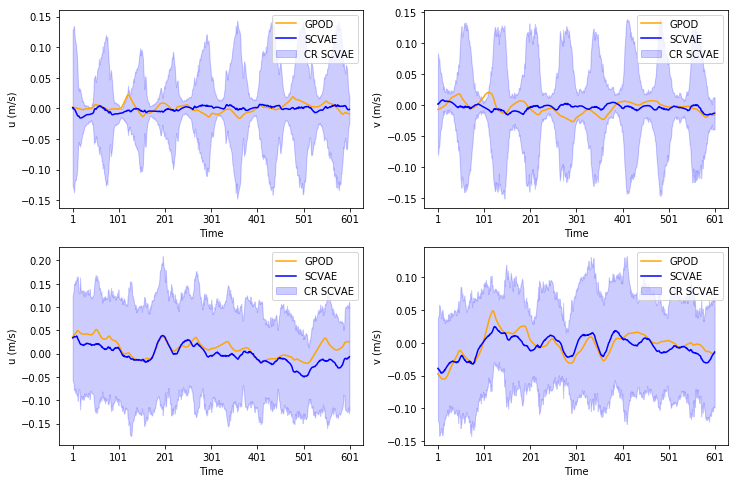}
	\captionof{figure}{The difference between the true and predicted estimate for the SCVAE (blue) and for the GPOD (orange) based on the $\lambda=0$ model and $Q_3$ measurement locations. The light blue shaded region represents the difference marginals, obtained from the confidence region in \cref{BOM_time_series}. \textbf{Upper panels:} The difference between the true and predicted estimate at grid point $(26,6)$ for $u$ (left) and $v$ (right), \textbf{Lower panels:} The difference between the true and predicted estimate at point $(4,1)$ for $u$ (left) and $v$ (right).}\label{BOM_error}
\end{figure}

\begin{figure}[H]
    \centering
		\includegraphics[width=0.70\linewidth]{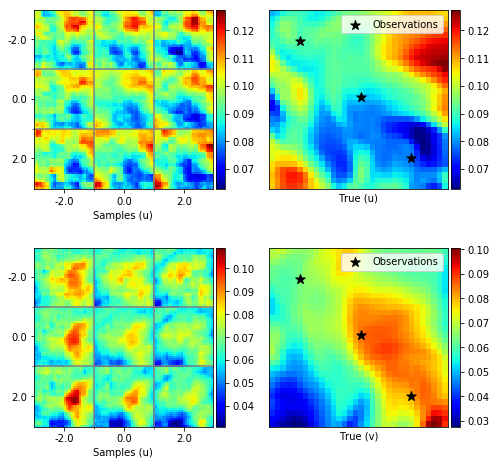}
		\captionof{figure}{The left panels shows 9 different generated velocity-field-snapshots for the $\uu$ and $\vv$ component for test sample number $1185$. The predictions are generated from the model with $\lambda=0$ and $Q_3$ measurement locations. We sample uniformly over the latent space and predicts with the decoder, given the measurements.}\label{z_space_sampling_v}
\end{figure}

In \cref{BOM_time_series} we plot the true solution and the predicted mean velocity \cref{eq:mean_and_cov} with the associated uncertainty, see \cref{eq:ConfRegion:2}, for two grid points. We plot only the first $600$ time steps for readability. The first grid point is $(26,6)$ and $(4,1).$ One location is approximately $5.1$ km from the nearest observation, and another one is about $16.1$ km away.
\cref{BOM_error} has to be viewed in context with \cref{BOM_time_series} and show the difference between the true and the predicted solutions with associated difference marginal in time for the two locations as in \cref{BOM_time_series}. 
Integrating over the latent space generates a posterior distribution of the reconstruction, as described in \cref{sec:posterior}. It is also possible to use the latent space to generate new statistically sound versions of $u$ and $v$. This is presented in \cref{z_space_sampling_v} where it is sampled uniformly over the 2 dimensional latent space $\z \sim \mathcal{N}(\bm{0},\bm{I})$ and the result shows how different variations can be created with the SCVAE model, given only the sparse measurements.    
These sampled velocities could be used for ensemble simulations when estimating uncertainty in a passive tracer transport, see e.g., \cite{oleynik2020optimal}. 

The SCVAE results are are compared with results of the GPOD method, see \cref{tab:comparison_BOM} and \cref{tab:comparison_BOM_div}. The tables show the errors as calculated in \cref{L2_error} and \cref{divergence_error_1} of the test data set for both sequential and random split.
For the sequential splitting, the SCVAE is better for $3$ measurement locations, while the GPOD method performs better for $9$ and $5$ locations.  From  \cref{Data_set_time_series_BOM}, we observe that test data set seems to arise from a different process than the train and validation data (especially for $v$). Thus, the SCVAE generalize worse than a simpler model such as the GPOD, \cite{model_selection}. For the $3$ location case, the number of components in the GPOD is not enough to compete with the SCVAE. \\

With random split on the train, test and validation data, we see that the SCVAE is significantly better than the GPOD. The training data and measurements represent the test data and test measurements better with random splitting. This highlights the importance of large data sets that cover as many outcomes as possible. Demanding that $\lambda > 0$ in \cref{obj_function_SCVAE} do not improve the result. The SCVAE-models with $\lambda = 0$ learns that the reconstructed representations should have low divergence without explicitly demanding it during optimization. However, as discussed in the 2D flow around cylinder experiment, demanding $\lambda>0$ seems to improve the conditioning of the optimization problem and give more consistent results.  
\begin{table}[ht]
    \centering
    \begin{tabular}{|c|c|c|c|c|c|} 
        \hline
         \multirow{2}{*}{Split} & \multirow{2}{*}{Regularization} & \multirow{2}{*}{Method} & \multicolumn{3}{|c|}{Measurement Locations} \\  \cline{4-6} 
          & & & 9 & 5 & 3 \\  \hline
         
         \multirow{4}{*}{Random} & \multirow{2}{*}{$\lambda=0$}
         & SCVAE & 0.1379 & 0.2097 & 0.2928   \\ \cline{3-6}
         & & GPOD & 0.3300 & 0.3822 & 0.4349  \\ \cline{2-6}
         & \multirow{2}{*}{$\lambda>0$}
         & SCVAE & 0.1403 & 0.2025 & 0.3016   \\ \cline{3-6}
         & & GPOD & 0.2971 & 0.3579 & 0.4039  \\ \cline{3-6} \hline
         
         \multirow{4}{*}{\makecell{Time \\ Dependent}} & \multirow{2}{*}{$\lambda=0$}
         & SCVAE & 0.3493 & 0.3913 & 0.4155  \\ \cline{3-6}
         & & GPOD & 0.3767 & 0.4031 & 0.4678  \\ \cline{2-6}
         & \multirow{2}{*}{$\lambda>0$}
         & SCVAE & 0.3527 & 0.3889 & 0.4141  \\ \cline{3-6}
         & & GPOD &0.3362 & 0.3695 & 0.4462  \\ \hline
    \end{tabular}
    \caption{Errors as calculated in \cref{L2_error} for the different methods, regularization techniques ($\lambda=0$ or $\lambda > 0$), split regimes and measurements}
    \label{tab:comparison_BOM}
\end{table}
\begin{table}[ht]
    \centering
    \begin{tabular}{|c|c|c|c|c|c|} 
        \hline
         \multirow{2}{*}{Split} & \multirow{2}{*}{Regularization} & \multirow{2}{*}{Method} & \multicolumn{3}{|c|}{Measurement Locations} \\  \cline{4-6} 
          & & & 9 & 5 & 3  \\  \hline
         
         \multirow{4}{*}{Random} & \multirow{2}{*}{$\lambda=0$}
         & SCVAE & 3.75e-05 & 3.62e-05 & 3.42e-05   \\ \cline{3-6}
         & & GPOD & 6.51e-05 & 5.88e-05 & 5.02e-05   \\ \cline{2-6}
         & \multirow{2}{*}{$\lambda>0$}
         & SCVAE &  3.60e-05 & 3.60e-05 & 3.13e-05  \\ \cline{3-6}
         & & GPOD & 6.23e-05 & 4.77e-05 & 4.14e-05  \\ \cline{3-6} \hline
         
         \multirow{4}{*}{\makecell{Time \\ Dependent}} & \multirow{2}{*}{$\lambda=0$}
         & SCVAE & 2.02e-05 & 1.80e-05 & 1.69e-05   \\ \cline{3-6}
         & & GPOD & 5.09e-05 & 4.03e-05 & 4.15e-05  \\ \cline{2-6}
         & \multirow{2}{*}{$\lambda>0$}
         & SCVAE & 2.05e-05 & 1.99e-05 & 1.85e-05  \\ \cline{3-6}
         & & GPOD & 4.39e-05 & 3.65e-05 & 2.92e-05  \\ \hline
    \end{tabular}
    \caption{Divergence errors as calculated in \cref{divergence_error_1} for the different methods, regularization techniques ($\lambda=0$ or $\lambda > 0$), split regimes and measurements. The true divergence of the test data is of order  $10^{-4}.$ }
    \label{tab:comparison_BOM_div}
\end{table}
\begin{figure}[H]
    \centering
 	    \includegraphics[width=0.75\linewidth]{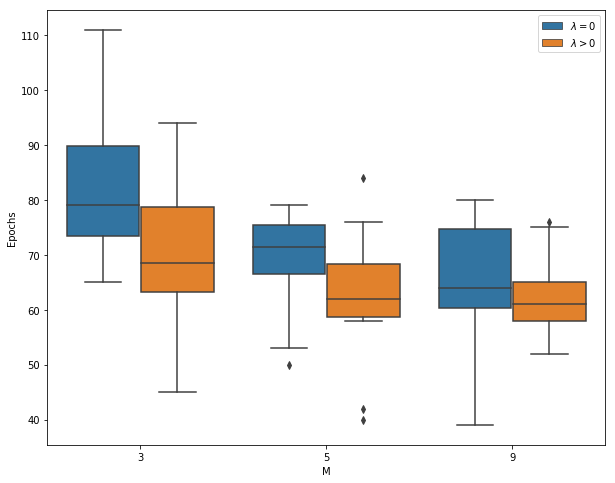}
 	    \captionof{figure}{The figure shows number of epochs and number of measurement locations. For each measurement configuration and regularization technique the model is optimized 10 times. The variation in the number of epochs for each measurement and regularization technique is due to different priors of the weights and mini-batch sampling.}\label{Epochs_per_measurement}
\end{figure}

\section{Discussion}\label{discussion}
We have presented the SCVAE method for efficient data reconstruction based on sparse observations. The derived objective functions for the network optimization  show that the encoding is independent of measurements. This allows for a simpler model structure with fewer model parameters than a CVAE and results in an optimization procedure that requires less computations.  \\

We have shown that the SCVAE is suitable for reconstruction of fluid flow. The method is showcased on two different data sets, velocity data from simulations of 2D flow around a cylinder, and bottom currents from the BOM. The fact that the fluids studied in the experiments are incompressible served as a motivation for adding an extra term to the objective function, see \cref{obj_function_SCVAE} with $\lambda>0$. \\

Our investigation of additional regularization showed that the mean reconstruction error over all models was lower with $\lambda>0$ compared to the model where $\lambda=0$, but the best reconstruction error was similar for $\lambda=0$ and $\lambda>0$. In \cref{Experiment} we optimized 10 models for every experiment, and chose the model that performed best on the validation data sets. With this approach we did not observe significant differences between optimizing with  $\lambda=0$ and $\lambda>0$. However, the reconstruction became less sensitive to the stochasticity involved in optimization (minibatch selection, network weights priors) when the regularization was used, see \cref{A_Motivating_Example}. \\ 

The SCVAE is a probabilistic model, which allows to make predictions, estimate their uncertainty, see \cref{sec:posterior},  and draw multiple samples from the predictive distribution, see \cref{z_space_sampling_v}. The last two properties make the SCVAE a useful method especially when the predictions are used in another application, i.e., ensemble simulation of tracer transport. Motivated by \cite{YILDIRIM2009160}, we compared the SCVAE predictions with the predictions of a modified GPOD method, see \cref{Appendix_C}. \\

Unlike the GPOD-method, a benefit with the SCVAE-method is that it scales well to larger data sets. Another aspect and as the experiments in \cref{Experiment} suggest, the GPOD seems more sensitive to the number of measurement locations than the SCVAE. On the other hand, the experiments suggested that GPOD is better than SCVAE with a larger number of measurement locations if the training data and the test data are too different, see BOM experiment with sequential splitting \cref{BOM_results}. Essentially the SCVAE overfit to the training data, and as a result performing poorly on the test data set. This fact shows the importance of training the the SCVAE on large data sets, which covers as many potential flow patterns as possible. Further, the results show that the GPOD is more sensitive to the measurement location choice than the SCVAE, see \cref{A_Motivating_Example}, and the GPOD-method is not expected to preform well on a complex flow with very few fixed measurement locations. \\

VAEs has been used for generating data in e.g. computer vision \cite{kingma2013auto}, and auto-encoders is a natural to use in reconstruction tasks \cite{ELMS2018}. Many reconstruction approaches, including the GPOD approach, first create a basis, then use the basis and minimize the error of the observations \cite{willcox2006unsteady, bui2004aerodynamic}. This makes the GPOD suitable for fast optimization of measurement locations that minimize the reconstruction error. On the other hand,  the SCVAE optimizes the basis function given the measurements, i.e. they are known and fixed. This makes it challenging to use the framework for optimizing sensor layout. But if the measurement locations are fixed and large amounts of training data are available, the SCVAE outperforms the GPOD for reconstruction. SCVAE optimize the latent representation and the neural network model parameters, variational and generative parameters, given the measurements. This ensures that the reconstruction is adapted to the specific configuration of measurements. \\

A limitation of our experiments is that we used only $100$ and $200$ samples and constructed the confidence region under further simplifying assumptions. The uncertainty estimate could be improved by increasing the sample size and better model for the confidence region. \\

Natural applications for the SCVAE are related to environmental data, where we often have sparse measurements. It is for example possible to optimize sensor layout to best possible detect unintentional discharges in the marine environment by using a simple transport model \cite{oleynik2020optimal}. Oleynik. et al. used deterministic flow fields to transport the contaminant and thus obtain a footprint of the leakage. SCVAE can be used to improve that method and efficiently generate probabilistic footprints of a discharges. This may be important as input to design, environmental risk assessments, and emergency preparedness plans. 

We have highlighted the SCVAE through the reconstruction of currents and flow field reconstruction, however, the SCVAE method is not limited to fluid flow problems. For instance, the same principles could be used in computer vision to generate new picture based on sparse pixel representations or in time series reconstruction. \\

A natural extension of the SCVAE is to set it up as a partially hidden Markov model. That is to predict the current state $p_\theta(\x_t|\m_t, \x_{t-1}),$ given the measurements and the reconstruction from the previous time step. This could potentially improve the reconstruction further.

\section*{Acknowledgements}
This work is part of the project ACTOM, funded through the ACT programme (Accelerating CCS Technologies, Horizon2020 Project No 294766). Financial contributions made from; The Research Council of Norway, (RCN), Norway, Netherlands Enterprise Agency (RVO), Netherlands, Department for Business, Energy \& Industrial Strategy (BEIS) together with extra funding from NERC and EPSRC research councils, United Kingdom, US-Department of Energy (US-DOE), USA. Kristian Gundersen has been supported by the Research Council of Norway, through the CLIMIT program (project 254711, BayMode) and the European Union Horizon 2020 research and innovation program under grant agreement 654462, STEMM-CCS. The authors would like to acknowledge NVIDIA Corporation for providing their GPUs in the academic GPU Grant Program.

\clearpage

\bibliographystyle{unsrt}
\clearpage
\bibliography{references}

\begin{thebibliography}{10}

\bibitem{brunton2015closed}
Steven~L Brunton and Bernd~R Noack.
\newblock Closed-loop turbulence control: Progress and challenges.
\newblock {\em Applied Mechanics Reviews}, 67(5), 2015.

\bibitem{kong2018application}
Lingxing Kong, Wei Wei, and Qingdong Yan.
\newblock Application of flow field decomposition and reconstruction in
  studying and modeling the characteristics of a cartridge valve.
\newblock {\em Engineering Applications of Computational Fluid Mechanics},
  12(1):385--396, 2018.

\bibitem{bolton2019applications}
Thomas Bolton and Laure Zanna.
\newblock Applications of deep learning to ocean data inference and subgrid
  parameterization.
\newblock {\em Journal of Advances in Modeling Earth Systems}, 11(1):376--399,
  2019.

\bibitem{venturi2004gappy}
Daniele Venturi and George~Em Karniadakis.
\newblock Gappy data and reconstruction procedures for flow past a cylinder.
\newblock {\em Journal of Fluid Mechanics}, 519:315, 2004.

\bibitem{callaham2018robust}
Jared Callaham, Kazuki Maeda, and Steven~L Brunton.
\newblock Robust flow field reconstruction from limited measurements via sparse
  representation.
\newblock {\em arXiv preprint arXiv:1810.06723}, 2018.

\bibitem{manohar2018data}
Krithika Manohar, Bingni~W Brunton, J~Nathan Kutz, and Steven~L Brunton.
\newblock Data-driven sparse sensor placement for reconstruction: Demonstrating
  the benefits of exploiting known patterns.
\newblock {\em IEEE Control Systems Magazine}, 38(3):63--86, 2018.

\bibitem{yeo2019data}
Kyongmin Yeo.
\newblock Data-driven reconstruction of nonlinear dynamics from sparse
  observation.
\newblock {\em Journal of Computational Physics}, 395:671--689, 2019.

\bibitem{oikonomou2018novel}
Panagiotis~D Oikonomou, Ayman~H Alzraiee, Christos~A Karavitis, and Reagan~M
  Waskom.
\newblock A novel framework for filling data gaps in groundwater level
  observations.
\newblock {\em Advances in Water Resources}, 119:111--124, 2018.

\bibitem{sirovich1987turbulence}
Lawrence Sirovich.
\newblock Turbulence and the dynamics of coherent structures. i. coherent
  structures.
\newblock {\em Quarterly of applied mathematics}, 45(3):561--571, 1987.

\bibitem{everson1995karhunen}
Richard Everson and Lawrence Sirovich.
\newblock Karhunen--loeve procedure for gappy data.
\newblock {\em JOSA A}, 12(8):1657--1664, 1995.

\bibitem{donoho2006compressed}
David~L Donoho.
\newblock Compressed sensing.
\newblock {\em IEEE Transactions on information theory}, 52(4):1289--1306,
  2006.

\bibitem{schmid2010dynamic}
Peter~J Schmid.
\newblock Dynamic mode decomposition of numerical and experimental data.
\newblock {\em Journal of fluid mechanics}, 656:5--28, 2010.

\bibitem{ELMS2018}
Al~Mamun S.~M. A., Lu~C., and Jayaraman B.
\newblock Extreme learning machines as encoders for sparse reconstruction.
\newblock {\em Fluids}, 3(4), 2018.

\bibitem{raissi2019physics}
Maziar Raissi, Paris Perdikaris, and George~E Karniadakis.
\newblock Physics-informed neural networks: A deep learning framework for
  solving forward and inverse problems involving nonlinear partial differential
  equations.
\newblock {\em Journal of Computational Physics}, 378:686--707, 2019.

\bibitem{raissi2018hidden}
Maziar Raissi, Alireza Yazdani, and George~Em Karniadakis.
\newblock Hidden fluid mechanics: A navier-stokes informed deep learning
  framework for assimilating flow visualization data.
\newblock {\em arXiv preprint arXiv:1808.04327}, 2018.

\bibitem{grover2019uncertainty}
Aditya Grover and Stefano Ermon.
\newblock Uncertainty autoencoders: Learning compressed representations via
  variational information maximization.
\newblock In {\em The 22nd International Conference on Artificial Intelligence
  and Statistics}, pages 2514--2524, 2019.

\bibitem{Rumelhart86_autoencoder}
D.~E. Rumelhart, G.~E. Hinton, and R.~J. Williams.
\newblock {\em Learning Internal Representations by Error Propagation}, page
  318–362.
\newblock MIT Press, Cambridge, MA, USA, 1986.

\bibitem{pearson1901_PCA}
Karl~Pearson F.R.S.
\newblock Liii. on lines and planes of closest fit to systems of points in
  space.
\newblock {\em The London, Edinburgh, and Dublin Philosophical Magazine and
  Journal of Science}, 2(11):559--572, 1901.

\bibitem{bourlard1988auto}
Herv{\'e} Bourlard and Yves Kamp.
\newblock Auto-association by multilayer perceptrons and singular value
  decomposition.
\newblock {\em Biological cybernetics}, 59(4-5):291--294, 1988.

\bibitem{kingma2013auto}
Diederik~P Kingma and Max Welling.
\newblock Auto-encoding variational bayes.
\newblock {\em arXiv preprint arXiv:1312.6114}, 2013.

\bibitem{sohn2015learning}
Kihyuk Sohn, Honglak Lee, and Xinchen Yan.
\newblock Learning structured output representation using deep conditional
  generative models.
\newblock In {\em Advances in neural information processing systems}, pages
  3483--3491, 2015.

\bibitem{MacKay92}
David J.~C. MacKay.
\newblock A practical bayesian framework for backpropagation networks.
\newblock {\em Neural Computation}, 4(3):448--472, 1992.

\bibitem{hoffman2013stochastic}
Matthew~D Hoffman, David~M Blei, Chong Wang, and John Paisley.
\newblock Stochastic variational inference.
\newblock {\em The Journal of Machine Learning Research}, 14(1):1303--1347,
  2013.

\bibitem{blei2017variational}
David~M Blei, Alp Kucukelbir, and Jon~D McAuliffe.
\newblock Variational inference: A review for statisticians.
\newblock {\em Journal of the American Statistical Association},
  112(518):859--877, 2017.

\bibitem{Halpern:2012hs}
Benjamin~S Halpern, Catherine Longo, Darren Hardy, Karen~L McLeod, Jameal~F
  Samhouri, Steven~K Katona, Kristin Kleisner, Sarah~E Lester, Jennifer
  O{\textquoteright}Leary, Marla Ranelletti, Andrew~A Rosenberg, Courtney
  Scarborough, Elizabeth~R Selig, Benjamin~D Best, Daniel~R Brumbaugh, F~Stuart
  Chapin, Larry~B Crowder, Kendra~L Daly, Scott~C Doney, Cristiane Elfes,
  Michael~J Fogarty, Steven~D Gaines, Kelsey~I Jacobsen, Leah~Bunce Karrer,
  Heather~M Leslie, Elizabeth Neeley, Daniel Pauly, Stephen Polasky, Bud Ris,
  Kevin St~Martin, Gregory~S Stone, U~Rashid Sumaila, and Dirk Zeller.
\newblock {An index to assess the health and benefits of the global ocean}.
\newblock {\em Nature}, 488(7413):615--620, August 2012.

\bibitem{DominguezTejo:2016dt}
Elianny Dom{\'\i}nguez-Tejo, Graciela Metternicht, Emma Johnston, and Luke
  Hedge.
\newblock {Marine Spatial Planning advancing the Ecosystem-Based Approach to
  coastal zone management: A review}.
\newblock {\em Marine Policy}, 72:115--130, October 2016.

\bibitem{drange2001ocean}
Helge Drange, Guttorm Alendal, and Ola~M Johannessen.
\newblock Ocean release of fossil fuel co2: A case study.
\newblock {\em Geophysical Research Letters}, 28(13):2637--2640, 2001.

\bibitem{BARSTOW1983211}
S.F. Barstow.
\newblock The ecology of langmuir circulation: A review.
\newblock {\em Marine Environmental Research}, 9(4):211 -- 236, 1983.

\bibitem{Ali:2011hd}
Alfatih Ali, {\O}yvind Thiem, and Jarle Berntsen.
\newblock {Numerical modelling of organic waste dispersion from fjord located
  fish farms}.
\newblock {\em Ocean Dynamics}, 61(7):977--989, apr 2011.

\bibitem{Law:2017}
Kara~Lavender Law.
\newblock Plastics in the marine environment.
\newblock {\em Annual Review of Marine Science}, 9(1):205--229, 2017.
\newblock PMID: 27620829.

\bibitem{Hylland:2015gt}
Ketil Hylland, Thierry Burgeot, Concepci{\'o}n Mart{\'\i}nez-G{\'o}mez, Thomas
  Lang, Craig~D Robinson, J{\"o}rundur Svavarsson, John~E Thain, A~Dick
  Vethaak, and Matthew~J Gubbins.
\newblock {How can we quantify impacts of contaminants in marine ecosystems?
  The ICON project}.
\newblock {\em Marine Environmental Research}, nov 2015.

\bibitem{Ali:2016go}
Alfatih Ali, H{\aa}vard~G Fr{\o}ysa, Helge Avlesen, and Guttorm Alendal.
\newblock {Simulating spatial and temporal varying {CO$_2$} signals from
  sources at the seafloor to help designing risk-based monitoring programs}.
\newblock {\em Journal Of Geophysical Research-Oceans}, 121(1):745--757,
  January 2016.

\bibitem{Blackford:2020}
Jerry Blackford, Guttorm Alendal, Helge Avlesen, Ashley Brereton, Pierre~W.
  Cazenave, Baixin Chen, Marius Dewar, Jason Holt, and Jack Phelps.
\newblock Impact and detectability of hypothetical ccs offshore seep scenarios
  as an aid to storage assurance and risk assessment.
\newblock {\em International Journal of Greenhouse Gas Control}, 95:102949,
  2020.

\bibitem{Hvidevold:2015}
Hilde~Kristine Hvidevold, Guttorm Alendal, Truls Johannessen, Alfatih Ali,
  Trond Mannseth, and Helge Avlesen.
\newblock {Layout of CCS monitoring infrastructure with highest probability of
  detecting a footprint of a {CO$_2$} leak in a varying marine environment}.
\newblock {\em International Journal of Greenhouse Gas Control}, 37:274--279,
  2015.

\bibitem{Hvidevold:2016cx}
Hilde~Kristine Hvidevold, Guttorm Alendal, Truls Johannessen, and Alfatih Ali.
\newblock {Survey strategies to quantify and optimize detecting probability of
  a {CO$_2$} seep in a varying marine environment}.
\newblock {\em Environmental Modelling {\&} Software}, 83:303--309, September
  2016.

\bibitem{Alendal:2017b}
Guttorm Alendal.
\newblock Cost efficient environmental survey paths for detecting continuous
  tracer discharges.
\newblock {\em Journal of Geophysical Research-Oceans}, 2017.
\newblock in press.

\bibitem{oleynik2020optimal}
Anna Oleynik, Maribel~I Garc{\'\i}a-Ib{\'a}{\~n}ez, Nello Blaser, Abdirahman
  Omar, and Guttorm Alendal.
\newblock Optimal sensors placement for detecting co2 discharges from unknown
  locations on the seafloor.
\newblock {\em International Journal of Greenhouse Gas Control}, 95:102951,
  2020.

\bibitem{gundersen2020binary}
Kristian Gundersen, Guttorm Alendal, Anna Oleynik, and Nello Blaser.
\newblock Binary time series classification with bayesian convolutional neural
  networks when monitoring for marine gas discharges.
\newblock {\em Algorithms}, 13(6):145, 2020.

\bibitem{willcox2006unsteady}
Karen Willcox.
\newblock Unsteady flow sensing and estimation via the gappy proper orthogonal
  decomposition.
\newblock {\em Computers \& fluids}, 35(2):208--226, 2006.

\bibitem{jo2019effective}
Taehyun Jo, Bonchan Koo, Hyunsoo Kim, Dohyung Lee, and Joon~Yong Yoon.
\newblock Effective sensor placement in a steam reformer using gappy proper
  orthogonal decomposition.
\newblock {\em Applied Thermal Engineering}, 154:419--432, 2019.

\bibitem{mifsud2019fusing}
Michael Mifsud, Alexander Vendl, Lars-Uwe Hansen, and Stefan G{\"o}rtz.
\newblock Fusing wind-tunnel measurements and cfd data using constrained gappy
  proper orthogonal decomposition.
\newblock {\em Aerospace Science and Technology}, 86:312--326, 2019.

\bibitem{callaham2019robust}
Jared~L Callaham, Kazuki Maeda, and Steven~L Brunton.
\newblock Robust flow reconstruction from limited measurements via sparse
  representation.
\newblock {\em Physical Review Fluids}, 4(10):103907, 2019.

\bibitem{weinkauf2010streak}
Tino Weinkauf and Holger Theisel.
\newblock Streak lines as tangent curves of a derived vector field.
\newblock {\em IEEE Transactions on Visualization and Computer Graphics},
  16(6):1225--1234, 2010.

\bibitem{gerrisflowsolver}
S.~Popinet.
\newblock Free computational fluid dynamics.
\newblock {\em ClusterWorld}, 2(6), 2004.

\bibitem{Proctor2014}
J.L. Proctor, S.L. Brunton, B.W. Brunton, and J.N. Kutz.
\newblock Exploiting sparsity and equation-free architectures in complex
  systems.
\newblock {\em Eur. Phys. J. Special Topics}, 223:2665–--2684, 2014.

\bibitem{YILDIRIM2009160}
B.~Yildirim, C.~Chryssostomidis, and G.E. Karniadakis.
\newblock Efficient sensor placement for ocean measurements using
  low-dimensional concepts.
\newblock {\em Ocean Modelling}, 27(3):160 -- 173, 2009.

\bibitem{vae_intro}
Diederik Kingma and Max Welling.
\newblock An introduction to variational autoencoders.
\newblock {\em Foundations and Trends® in Machine Learning}, 12:307--392, 01
  2019.

\bibitem{gregor2015draw}
Karol Gregor, Ivo Danihelka, Alex Graves, Danilo~Jimenez Rezende, and Daan
  Wierstra.
\newblock Draw: A recurrent neural network for image generation.
\newblock {\em arXiv preprint arXiv:1502.04623}, 2015.

\bibitem{bowman2015generating}
Samuel~R Bowman, Luke Vilnis, Oriol Vinyals, Andrew~M Dai, Rafal Jozefowicz,
  and Samy Bengio.
\newblock Generating sentences from a continuous space.
\newblock {\em arXiv preprint arXiv:1511.06349}, 2015.

\bibitem{kullback1951information}
Solomon Kullback and Richard~A Leibler.
\newblock On information and sufficiency.
\newblock {\em The annals of mathematical statistics}, 22(1):79--86, 1951.

\bibitem{higgins2016beta}
Irina Higgins, Lo{\"i}c Matthey, Arka Pal, Christopher Burgess, Xavier Glorot,
  Matthew~M Botvinick, Shakir Mohamed, and Alexander Lerchner.
\newblock beta-{VAE}: Learning basic visual concepts with a constrained
  variational framework.
\newblock In {\em ICLR}, 2017.

\bibitem{kuhn2014nonlinear}
Harold~W Kuhn and Albert~W Tucker.
\newblock Nonlinear programming.
\newblock In {\em Traces and emergence of nonlinear programming}, pages
  247--258. Springer, 2014.

\bibitem{karush1939minima}
William Karush.
\newblock Minima of functions of several variables with inequalities as side
  constraints.
\newblock {\em M. Sc. Dissertation. Dept. of Mathematics, Univ. of Chicago},
  1939.

\bibitem{kiefer1952stochastic}
Jack Kiefer, Jacob Wolfowitz, et~al.
\newblock Stochastic estimation of the maximum of a regression function.
\newblock {\em The Annals of Mathematical Statistics}, 23(3):462--466, 1952.

\bibitem{robbins1951stochastic}
Herbert Robbins and Sutton Monro.
\newblock A stochastic approximation method.
\newblock {\em The annals of mathematical statistics}, pages 400--407, 1951.

\bibitem{berntsen2000users}
Jarle Berntsen.
\newblock Users guide for a modesplit $\sigma$-coordinate numerical ocean
  model.
\newblock {\em Department of Applied Mathematics, University of Bergen, Tech.
  Rep}, 135:48, 2000.

\bibitem{lecun1998gradient}
Yann LeCun, L{\'e}on Bottou, Yoshua Bengio, and Patrick Haffner.
\newblock Gradient-based learning applied to document recognition.
\newblock {\em Proceedings of the IEEE}, 86(11):2278--2324, 1998.

\bibitem{krizhevsky2012imagenet}
Alex Krizhevsky, Ilya Sutskever, and Geoffrey~E Hinton.
\newblock Imagenet classification with deep convolutional neural networks.
\newblock In {\em Advances in neural information processing systems}, pages
  1097--1105, 2012.

\bibitem{noh2015learning}
Hyeonwoo Noh, Seunghoon Hong, and Bohyung Han.
\newblock Learning deconvolution network for semantic segmentation.
\newblock In {\em Proceedings of the IEEE international conference on computer
  vision}, pages 1520--1528, 2015.

\bibitem{lecun1989backpropagation}
Yann LeCun, Bernhard Boser, John~S Denker, Donnie Henderson, Richard~E Howard,
  Wayne Hubbard, and Lawrence~D Jackel.
\newblock Backpropagation applied to handwritten zip code recognition.
\newblock {\em Neural computation}, 1(4):541--551, 1989.

\bibitem{kingma2014adam}
Diederik~P Kingma and Jimmy Ba.
\newblock Adam: A method for stochastic optimization.
\newblock {\em arXiv preprint arXiv:1412.6980}, 2014.

\bibitem{heydari2019softadapt}
A~Ali Heydari, Craig~A Thompson, and Asif Mehmood.
\newblock Softadapt: Techniques for adaptive loss weighting of neural networks
  with multi-part loss functions.
\newblock {\em arXiv preprint arXiv:1912.12355}, 2019.

\bibitem{model_selection}
Kenneth Burnham and David Anderson.
\newblock Model selection and multimodel inference.
\newblock {\em A Practical Information-theoretic Approach}, 01 2004.

\bibitem{bui2004aerodynamic}
Tan Bui-Thanh, Murali Damodaran, and Karen Willcox.
\newblock Aerodynamic data reconstruction and inverse design using proper
  orthogonal decomposition.
\newblock {\em AIAA journal}, 42(8):1505--1516, 2004.

\bibitem{chollet2015keras}
Fran\c{c}ois Chollet et~al.
\newblock Keras.
\newblock \url{https://keras.io}, 2015.

\end{thebibliography}

\clearpage
\appendix

\section{GPOD method with divergence regularization}\label{Appendix_C}
Let the vectors $\x^{(i)}$ in $\X$ be organized as a snapshots matrix $\X_h \in \R^{2N \times K}$. 
Here we consider the latent space be given by the $r$ principal components of the matrix $\X_h$, assuming $r << N.$  
Thus, $\X_h \approx \Phi A_h$ where $\Phi$ is $2N \times r$ matrix of principal components, and $A_h= \Phi^+ X_h$ is $r \times K$ representation of $X_h.$

Let $\x^{*}  \in \R^{2N} $ be an unknown state that we need to reconstruct from $M$ spatial  measurements $\m^*=C \x^{*}$. 
We assume that there is ${\bm a}$ such that $\x^*= \bm{\Phi a}$. Therefore we search for a solution of $\bm{C \Phi a}=\m^{*}.$
Even if the system $\bm{C \Phi a}=\m^{*}$  is overdetermined, the matrix $C \Phi$ could be ill-conditioned or rank-deficient. However, since the number of sensors is usually small, $2M<r,$ the system  is underdetermined and regularization is required.

Assuming that the flow is incompressible, the natural regularization is to penalize the divergence error of a solution. That is, we solve
 
\begin{equation}
\label{eq:Aopt}
{\bm a^{*}}= argmin_{\bm a} \| \bm{ C \Phi a}-\m^{*}\|^2_2 + \lambda \| L_{div} \bm{\Phi a}\|^2_2 
\end{equation}
where $L_{div}:\R^{2N} \to \R^{N} $ is a linear operator approximating the divergence, and $\lambda >0$ is a regularization constant.
Finally, we decode  $\x^{*}$ from the measurements $\m^{*}$ as $\x^{*}=\bm{\Phi  a^{*}}.$ 
\clearpage
\section{Scaling of a data}\label{Appendix_D}

Let $\cT_{train}$ contains the times $t_{l_i},$ $i=1,...,n$ corresponding to the training data.
We define 
$$
u_{\max}
=\max\limits_{p, t} u(p, t) \quad \mbox{ and } 
 \quad u_{\min}=\min\limits_{p, t} u(p, t), \quad 
 $$
and
$$
v_{\max}=\max \limits_{p, t } v(p, t) \quad \mbox{ and } \quad
v_{\min}=\min \limits_{p, t} v(p, t), 
$$
as the largest and smallest values of $u$
and $v$ on $\cP$ and $\cT_{train}.$
Then, the middle points are given as
$$
u_c=\frac{u_{\max} +u_{\min} }{2}, \quad
v_c=\frac{v_{\max} +v_{\min} }{2},
$$
and the half lengths as
$$
d_u=\frac{u_{\max} -u_{\min} }{2}, \quad
d_v=\frac{v_{\max} - v_{\min} }{2}.
$$

Then the whole data is scaled as
$$
\tilde{u}=\frac{u-u_c}{d_u} \quad \tilde{v}=\frac{v-v_c}{d_v},
$$
and the divergence operator $L_{div}$ scaled accordingly.

After the optimization is completed, the data is scaled back, i.e.,
$$
u=d_u\tilde{u}+u_c \quad v=d_v\tilde{v}+v_c.
$$

The relative errors \cref{L2_error} are calculated on the scaled data. The divergence error \cref{divergence_error_1} is unaffected by the scaling.

\clearpage
\section{Details on the Experiments}\label{Appendix_A}
We use Keras \cite{chollet2015keras} in the implementation of the SCVAE for all experiments. In this Appendix we present details on the architecture of the decoders and encoders for the different experiments. We have optimized the SCVAE models with different number of measurements. That is, the shape of the input layer to the decoder will be dependent on the measurements (sensor-input layer). Here we present details on the architecture of the encoders and decoders with largest number of measurements for SCVAE models for both experiments. There is one extra dimension in the figures showing the encoders and decoders. This dimension is here one, but the framework is implemented to allow for more dimensions in time.    

\subsection{Encoder for 2D Flow Around Cylinder Data Experiment}\label{Appendix_A.1}
\begin{figure}[h]
	\centering
		\includegraphics[width=0.70\linewidth]{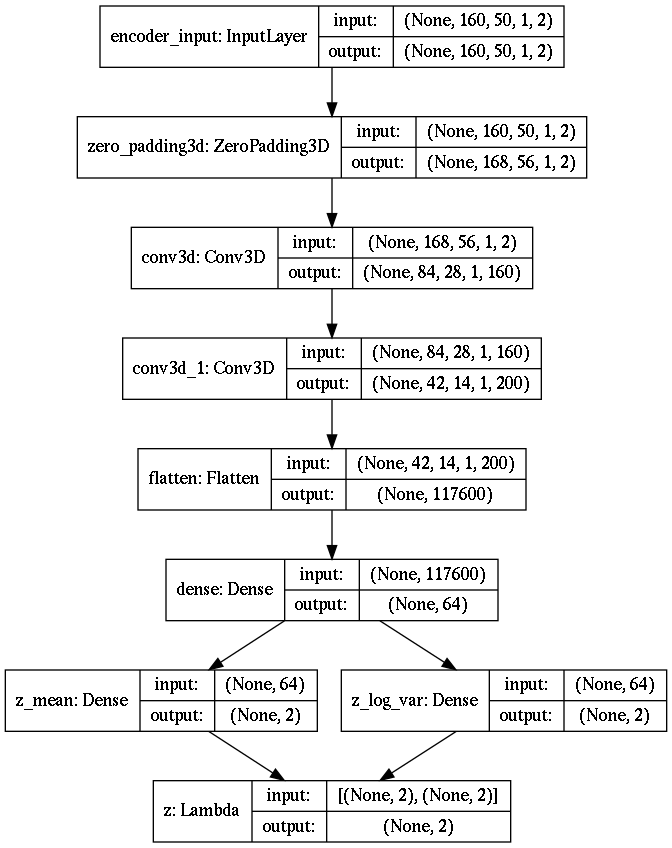}
		\captionof{figure}{}\label{Encoder_CW}
\end{figure}

\newpage
\subsection{Decoder for 2D Flow Around Cylinder Experiment}\label{Appendix_A.2}
\begin{figure}[h]
	\centering
		\includegraphics[width=0.70\linewidth]{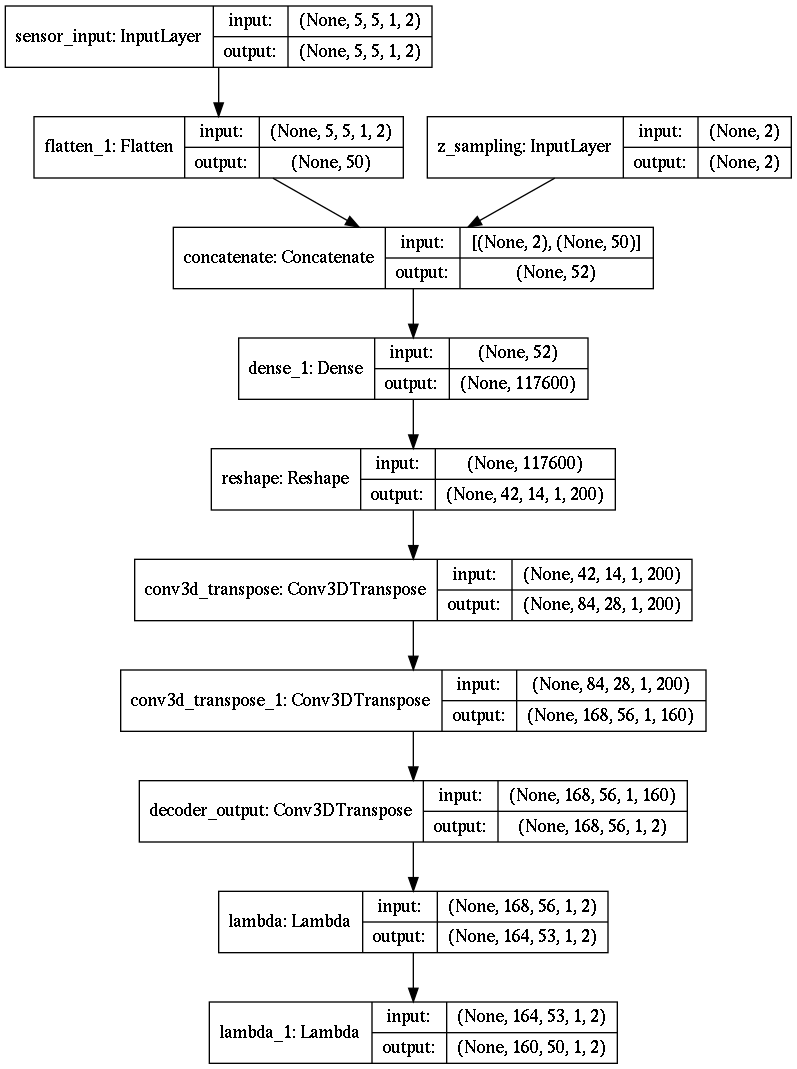}
		\captionof{figure}{}\label{decoder_CW}
\end{figure} 

\newpage
\subsection{Encoder for BOM Data Experiment}\label{Appendix_A.3}
\begin{figure}[h]
	\centering
		\includegraphics[width=0.70\linewidth]{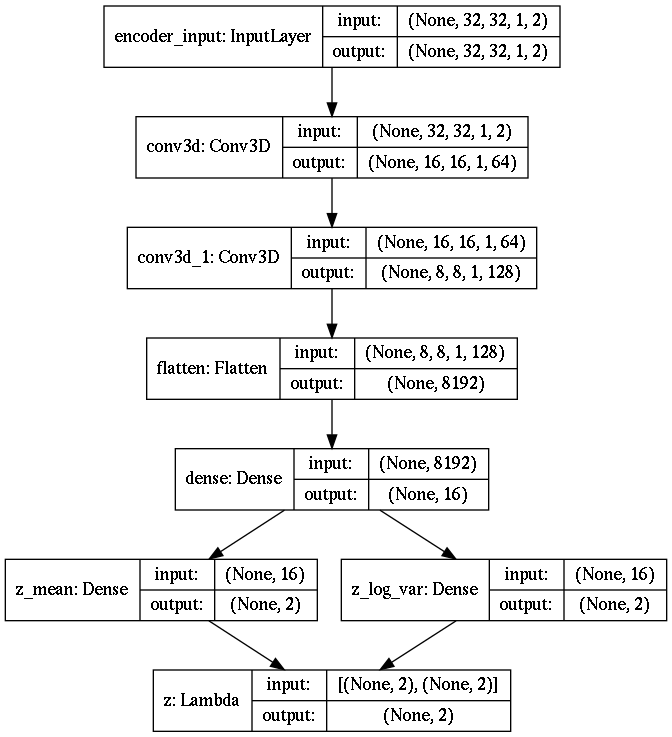}
		\captionof{figure}{}\label{Encoder_BOM}
\end{figure}

\newpage
\subsection{Decoder for BOM Data Experiment}\label{Appendix_A.4}
\begin{figure}[h]
	\centering
		\includegraphics[width=0.70\linewidth]{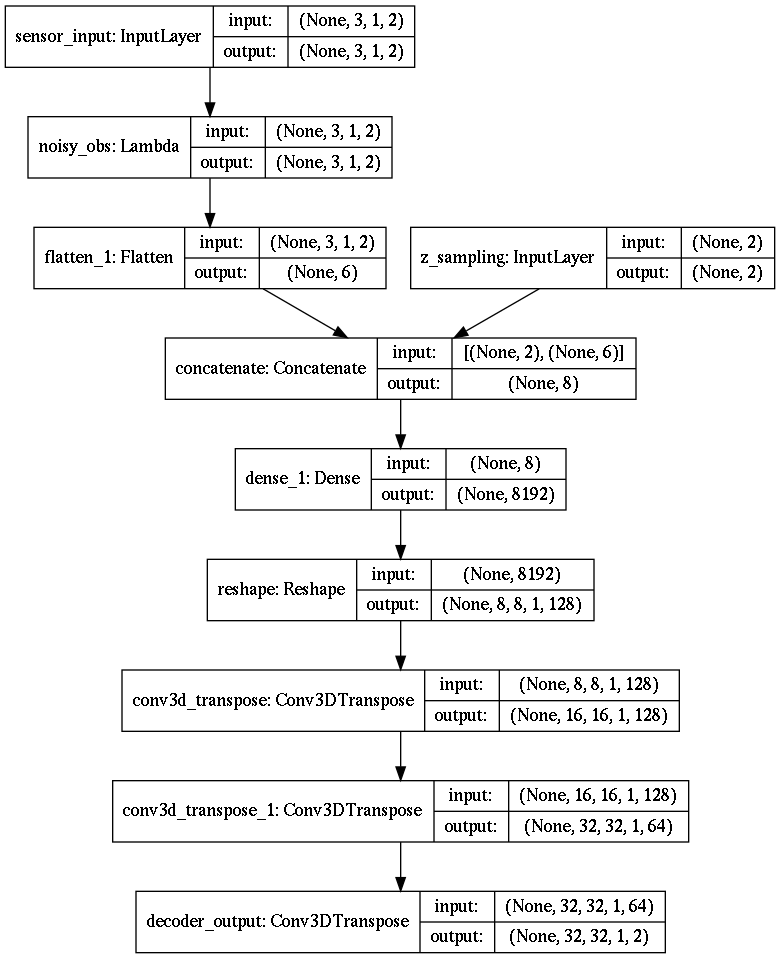}
		\captionof{figure}{}\label{decoder_BOM}
\end{figure}

\clearpage

\end{document}